\documentclass[Afour, sageh, times]{sagej}
\usepackage{times}
\usepackage{moreverb,url}
\usepackage{xcolor}
\usepackage{soul}

\usepackage[colorlinks,bookmarksopen,bookmarksnumbered,citecolor=red,urlcolor=red]{hyperref}

\usepackage{graphics}           
\usepackage{times}              
\usepackage{amsmath}            
\usepackage{amssymb}            
\usepackage{graphicx}
\usepackage{algorithm}
\usepackage[noend]{algpseudocode}
\usepackage{booktabs}
\usepackage{color}
\usepackage{listings}
\usepackage{subfiles}
\usepackage{subcaption}
\usepackage{multirow}
\usepackage{multicol}
\usepackage{makecell}
\usepackage{array}
\definecolor{instructioncolor}{rgb}{.5,.5,.5}

\usepackage[font=small]{caption}

\def\secref#1{Sec.~\ref{#1}}
\def\figref#1{Fig.~\ref{#1}}
\def\tabref#1{Tab.~\ref{#1}}
\def\eqref#1{Eq.~(\ref{#1})}

\makeatletter
\usepackage{xspace}
\DeclareRobustCommand\onedot{\futurelet\@let@token\@onedot}
\def\@onedot{\ifx\@let@token.\else.\null\fi\xspace}
\def\eg{e.g\onedot} 
\def\ie{i.e\onedot}

\def\wrt{w.r.t\onedot} 

\makeatother

\newcommand{\tbf}[1]{\textbf{#1}}
\newcommand{\tul}[1]{\underline{#1}}

\usepackage{array}
\newcolumntype{L}[1]{>{\raggedright\let\newline\\\arraybackslash\hspace{0pt}}m{#1}}
\newcolumntype{C}[1]{>{\centering\let\newline\\\arraybackslash\hspace{0pt}}m{#1}}
\newcolumntype{R}[1]{>{\raggedleft\let\newline\\\arraybackslash\hspace{0pt}}m{#1}}

\def\kiss{KISS-ICP}
\def\lidar{LiDAR}
\def\lidars{LiDARs}
\def\slam{SLAM}

\def\scancontext{Scan Context}
\def\solid{SOLiD}
\def\logg3d{LoGG3D}
\def\bevplace{BEVPlace++}

\def\nclt{NCLT}
\def\mulran{MulRan}
\def\helipr{HeLiPR}
\def\car{IPB-Car}
\def\backpack{IPB-Backpack}

\def\aeva{Aeva Aeries II}
\def\livox{Livox Avia}
\def\ouster{Ouster OS2-128}
\def\velodyne{Velodyne HDL-32E}

\def\sota{state-of-the-art}
\def\pr{precision-recall}
\def\fov{FoV}
\def\bev{BEV}
\def\map{local map}
\def\maps{local maps}
\def\ground{ground plane}
\def\xy{xy-plane}
\def\pg{pose-graph}

\graphicspath{
	{pics/}
}

\makeatletter
\def\input@path{
    {pics/}
    {tables/}
}
\makeatother

\def\argmin{\mathop{\rm argmin}}

\newcommand{\SE}[1]{\mathbb{SE}\,(#1)}
\newcommand{\SO}[1]{\mathbb{SO}\,(#1)}

\newcommand{\RR}{\mathbb{R}}
\newcommand{\NN}{\mathbb{N}}

\newcommand{\norm}[1]{\lVert#1\lVert}

\renewcommand{\b}[1]{\mbox{\boldmath$#1$}}

\renewcommand{\v}[1]{{\b #1}} 

\newcommand{\m}[1]{{\mbox{{\sffamily\slshape{#1\/}}}}}

\newcommand{\mq}[1]{{\mbox{{\sffamily{#1}}}}}

\newcommand{\tr}[0]{\sf T}              

\newcommand{\inv}{^{-1}}

\newcommand{\degrees}{{\mbox{$^\circ$}}}

\newcommand{\T}[2]{{}^{#1}{\mq{T}}_{#2}}
\newcommand{\pts}[3]{{#1}_{#2}^{#3}}

\newcommand{\zvector}[2]
  {    \left[
          \begin{array}{c}
            {#1} \\ {#2}
        \end{array}
       \right] }

\newcommand{\zvectort}[2]
  {    \left[
            {#1} \; {#2}
       \right] }

\newcommand{\dvectort}[3]
  {    \left[
            {#1} \; {#2}\; {#3}
       \right] }

\newcommand{\svectort}[6]
  {    \left[
            {#1} \; {#2}\; {#3}\; {#4}\; {#5}\; {#6}
       \right] }

\newcommand{\bg}{\b g}

\newcommand{\mD}{\m D}

\newcommand{\mG}{\m G}

\newcommand{\mI}{\m I}
\newcommand{\mJ}{\m J}

\newcommand{\mM}{\m M}
\newcommand{\mN}{\m N}

\newcommand{\mP}{\m P}

\newcommand{\mS}{\m S}

\begin{document}
\runninghead{Gupta et al.}

\title{\LARGE \bf Efficiently Closing Loops in LiDAR-Based SLAM Using Point Cloud Density Maps}

\author{Saurabh Gupta\affilnum{1} and Tiziano Guadagnino\affilnum{1} and Benedikt Mersch\affilnum{1} and Niklas Trekel\affilnum{1} and Meher V. R. Malladi\affilnum{1} and Cyrill Stachniss\affilnum{1,2}}

\affiliation{\affilnum{1}University of Bonn, Center for Robotics\\
\affilnum{2}Lamarr Institute for Machine Learning and Artificial Intelligence
}

\corrauth{Saurabh Gupta,
University of Bonn,
Center for Robotics,
Germany.}

\email{sgupta2@uni-bonn.de}

\begin{abstract}
	Consistent maps are key for most autonomous mobile robots, and they often use SLAM approaches to build such maps. Loop closures via place recognition help to maintain accurate pose estimates by mitigating global drift, and are thus key for realizing an effective SLAM system. This paper presents a robust loop closure detection pipeline for outdoor SLAM with LiDAR-equipped robots. Our method handles various LiDAR sensors with different scanning patterns, fields of view, and resolutions. It generates local maps from LiDAR scans and aligns them using a ground alignment module to handle both planar and non-planar motion of the LiDAR, ensuring applicability across platforms. The method uses density-preserving bird's-eye-view projections of these local maps and extracts ORB feature descriptors for place recognition. It stores the feature descriptors in a binary search tree for efficient retrieval, and self-similarity pruning addresses perceptual aliasing in repetitive environments. Extensive experiments on public and self-recorded datasets demonstrate accurate loop closure detection, long-term localization, and cross-platform multi-map alignment, agnostic to the LiDAR scanning patterns, fields of view, and motion profiles. We provide the code for our pipeline as open-source software at \url{https://github.com/PRBonn/MapClosures}.
\end{abstract}

\maketitle

\footnote{Accepted at the International Journal of Robotics Research, April 2026.}
\begin{figure*}
	\fontsize{10}{10}\selectfont
	\def\svgwidth{0.99\linewidth}
\begingroup%
  \makeatletter%
  \providecommand\color[2][]{%
    \errmessage{(Inkscape) Color is used for the text in Inkscape, but the package 'color.sty' is not loaded}%
    \renewcommand\color[2][]{}%
  }%
  \providecommand\transparent[1]{%
    \errmessage{(Inkscape) Transparency is used (non-zero) for the text in Inkscape, but the package 'transparent.sty' is not loaded}%
    \renewcommand\transparent[1]{}%
  }%
  \providecommand\rotatebox[2]{#2}%
  \newcommand*\fsize{\dimexpr\f@size pt\relax}%
  \newcommand*\lineheight[1]{\fontsize{\fsize}{#1\fsize}\selectfont}%
  \ifx\svgwidth\undefined%
    \setlength{\unitlength}{591.87264029bp}%
    \ifx\svgscale\undefined%
      \relax%
    \else%
      \setlength{\unitlength}{\unitlength * \real{\svgscale}}%
    \fi%
  \else%
    \setlength{\unitlength}{\svgwidth}%
  \fi%
  \global\let\svgwidth\undefined%
  \global\let\svgscale\undefined%
  \makeatother%
  \begin{picture}(1,0.34201318)%
    \lineheight{1}%
    \setlength\tabcolsep{0pt}%
    \put(0,0){\includegraphics[width=\unitlength,page=1]{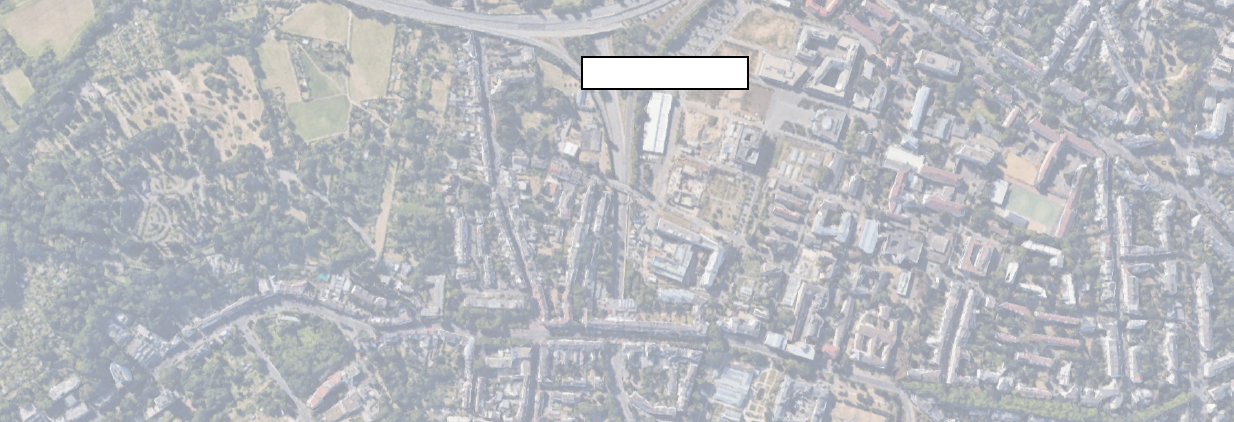}}%
    \put(0.53920504,0.27723037){\color[rgb]{0.2,0.4,0.6}\makebox(0,0)[t]{\lineheight{1.25}\smash{\begin{tabular}[t]{c}\textbf{\backpack{}}\end{tabular}}}}%
    \put(0,0){\includegraphics[width=\unitlength,page=2]{figure_1.pdf}}%
    \put(0.72726841,0.01679307){\color[rgb]{0.60392157,0,0}\makebox(0,0)[t]{\lineheight{1.25}\smash{\begin{tabular}[t]{c}\textbf{\car{}}\end{tabular}}}}%
    \put(0,0){\includegraphics[width=\unitlength,page=3]{figure_1.pdf}}%
    \put(0.51940558,0.01832472){\makebox(0,0)[t]{\lineheight{1.25}\smash{\begin{tabular}[t]{c}loop closure\end{tabular}}}}%
    \put(0,0){\includegraphics[width=\unitlength,page=4]{figure_1.pdf}}%
    \put(0.70917435,0.27788901){\makebox(0,0)[t]{\lineheight{1.25}\smash{\begin{tabular}[t]{c}loop closure\end{tabular}}}}%
    \put(0,0){\includegraphics[width=\unitlength,page=5]{figure_1.pdf}}%
  \end{picture}%
\endgroup%

	\captionof{figure}{An example of loop closures detected between two sequences recorded with different \lidar{} sensor platforms with a revisit interval of two weeks. In blue is the \backpack{} sequence, recorded on the campus of the University of Bonn with a Hesai Pandar-128 \lidar{} mounted on a backpack. In red is the \car{} sequence, recorded in the city of Bonn with an Ouster OS1-128 \lidar{}. Both trajectories were individually optimized through a \pg{} with in-session loop closure constraints obtained from our pipeline. We used our loop closure pipeline to detect inter-session loop closures between these two sequences, which have little overlap, and aligned the two trajectories using the multi-session loop closure constraints. The overlapping areas are highlighted by small rectangles, and the corresponding loop closure from these areas is shown in the enlarged rectangles, respectively.}\label{fig:1}
	\vspace{-5pt}
\end{figure*}%

\section{Introduction}\label{sec:intro}

Mobile robots must navigate their surroundings safely and efficiently. They need to know their location within the environment to successfully navigate to a desired place or explore new areas. Accurate ego-motion estimation helps robots generate accurate maps of the environment, which they can use for navigation. Traditionally, robots often localize themselves using data acquired through exteroceptive sensors like cameras and laser range sensors~(\lidar{}), proprioceptive sensors like wheel odometers and IMUs, or a combination. Depending on the application and availability, outdoor systems often also exploit GNSS for global positioning.

\lidar{} sensors are frequently used in robotics due to their accurate and dense 3D range data. Many previous studies have advanced the field of sequential pose estimation using \lidar{} sensors~\citep{vizzo2023ral,guadagnino2022ral,dellenbach2022icra,ferrari2024ral,fontana2016iros}. Such sequential pose estimation alone, also called sensor odometry, suffers from drifting pose estimates over time due to inherent noise in the robot motion and sensor data, dynamics in the environment, and non-trivial data association problems.

We can compensate for such drift and improve pose estimation by recognizing places previously visited by the robot. This task is referred to as loop closing. It allows the use of geometric information from the \lidar{} observations at revisited locations to correct the pose drift, \eg{}, through \pg{} optimization. Robust loop closure detection is paramount in simultaneous localization and mapping~(\slam{}) systems. Robots must recognize that they have returned to a previously visited place to close a loop.

Place recognition is a key sub-task within loop closure detection, which performs data associations between the robot's current view and a database of previously seen places. Generating a database of places is a non-trivial task that requires crafting an as unique as possible description of the environment, invariant to changes in viewpoint. Also, such a database should filter dynamic objects appearing in the sensor data to achieve robust place recognition across long time intervals. Perceptual aliasing is another challenge, as distinct places with similar structures can confuse place recognition algorithms. This could lead to false-positive loop closures that can adversely impact global SLAM estimates~\citep{blanco2013robotica,bailey2006ram,ramos2007icra}. Furthermore, to perform place recognition for loop closing in a \slam{} system, the database should encode the scene's geometry to allow for relative pose estimation between different revisit viewpoints for the \pg{} optimization.

Global feature descriptors computed over the entire point clouds are easy to store in a database and allow for quick retrieval of revisit candidates~\citep{kim2021tro,angelina2018cvpr,lu2022iros,xu2023tro,luo2025tro}. However, such methods often do not provide an initial geometric alignment between the detected loop closures and require an additional point cloud registration step to perform such an alignment. In contrast, feature descriptors computed over local 3D patches within a point cloud can aid the geometric alignment of revisited locations~\citep{blanco2013robotica,yang2017pr,steder2011iros}. However, they require a nuanced approach to store the feature descriptors efficiently in a database.

Due to the algorithmic complexity of processing 3D point clouds to obtain local or global feature descriptors, several methods perform a bird's-eye-view~(\bev{}) projection~\citep{kim2021tro,li2021icra-ligl,luo2023iccv} or a cylindrical projection~\citep{chen2020rss,steder2010icra,ma2022ral} of the point clouds. This results in a 2D representation of the point cloud data, allowing faster feature detection and matching for online loop closure detection in \slam{}.

The main contribution of this article is a robust loop closure detection pipeline that works with various \lidar{} sensors having different motion profiles, invariant to their scanning pattern, field of view~(\fov{}), and resolution. We generate \maps{} by aggregating consecutive scans and creating a density-preserving \bev{} projection of the \maps{} as an intermediate 2D representation for computing local features describing the structural information in the 3D \maps{}. Since the \ground{}s can be consistently identified in outdoor environments during revisits, we use them as a reference plane to ensure consistent \bev{} projections across revisits. We propose a ground alignment module to identify such a \ground{} in each \map{} and transform the \map{} such that this \ground{} coincides with the local \xy{} of the reference frame. This simplifies the \bev{} projection and restricts the loop closure alignment to two dimensions on the common \ground{} and makes our approach applicable to various motion profiles of \lidar{} sensors. We store binary ORB feature descriptors from the \bev{} projection into a Hamming distance embedding binary search tree~(HBST). We design our approach to be robust against the effects of scene similarity, also known as perceptual aliasing, by performing a self-similarity pruning of the feature descriptors before inserting them in the database. Using a Hamming distance metric, we obtain loop closure candidates by matching feature descriptors from query \maps{} against this database. Our subsequent random sample consensus~(RANSAC) based geometric validation step provides loop closures along with a 2D alignment between \bev{} projections of the \maps{}. When combined with the ground alignment estimate of each \map{}, we obtain a complete 3D estimate of the global alignment between the \maps{}. We provide an extensive experimental evaluation on multiple datasets, with sequences recorded with a wide range of \lidar{} sensors mounted on different mobile platforms, testing the accuracy and robustness of our pipeline in challenging scenarios.

In sum, we make six key claims, which we support with the paper and our experimental evaluation. Our approach
\begin{enumerate}
	\item detects loop closures between \maps{} generated from various \lidar{} sensors with different scanning patterns, \fov{}s, and resolutions;
	\item performs multi-session loop closure detection and alignment with long-term revisits;
	\item works with handheld platforms having non-planar motion in the \lidar{} sensor frame;
	\item is robust against perceptual aliasing in environments with repetitive structures;
	\item provides a complete 3D rigid-body transform to align the detected loop closures;
	\item detects loop closures between sequences having minimal overlap, recorded with different \lidar{} sensor platforms, enabling cross-platform multi-map alignment.
\end{enumerate}

This article extends our previous conference paper~\citep{gupta2024icra}, which proposed loop closure detection using local maps by detecting local feature descriptors from their BEV projections and generating a binary search tree database. It extends the earlier work in the following critical ways: (1) relaxation of the planar-motion assumption through a new ground alignment module; (2) a complete 3D alignment estimate of the detected loop closures; (3) a pruning strategy on the ORB feature descriptors computed from the BEV density images to mitigate the issues with perceptual aliasing; (4) an extensive experimental evaluation on multiple datasets, with sequences recorded with a wide range of \lidar{} sensors mounted on different mobile platforms, testing the accuracy and robustness of our pipeline in challenging scenarios.

The open-source implementation of our previous approach and the proposed approach are available at \url{https://github.com/PRBonn/MapClosures}.

\section{Related Work}\label{sec:related}

\cite{zhang2023acmcs} provide a thorough literature review on place recognition for loop closures in 3D \lidar{} SLAM and~\cite{yin2024ijcv} on \lidar{}-based global localization. This section briefly overviews key approaches for \lidar{}-based loop closure detection in \slam{}.

The most straightforward approach to loop closure detection in \slam{} is using the individual scans and their poses obtained from odometry. The similarity of poses between non-consecutive scans within a search radius is used to detect loop closures~\citep{rottmann2019ecmr,shan2020iros}. S4-SLAM~\citep{zhou2021ar} and the approach by~\cite{mendes2016ssrr} use the odometry information to compute the overlap between point clouds to check if they are recorded from the same location.~\cite{chen2020rss} use convolutional neural networks to calculate the overlap between the range image representations of point clouds to detect loops. These methods primarily perform a place retrieval task, often requiring a subsequent point cloud registration step to validate the loop closures. Furthermore, they can be sensitive to the magnitude of drift present in the odometry pose estimates, requiring a search radius proportional to the drift.
\subsection{Methods Using 3D Features From Point Clouds}
Place recognition has been widely studied in the context of camera images~\citep{cummins2008ijrr,galvez2012tro,vysotska2019ral,vysotska2016ral,mur-artal2015tro}. Consequently, much research on \lidar{} place recognition for loop closing has been motivated by approaches in the camera vision domain. In particular, many approaches have extended the idea of local feature descriptors to 3D point clouds~\citep{rusu2009icra,salti2014cviu,johnson1999pami}, discretizing the neighborhood around 3D features into a geometrical grid. They compute a descriptor from the points in this neighborhood based on their height~\citep{bosse2013icra}, density~\citep{tombari2010acmw,frome2004eccv}, or distance and angle~\citep{rusu2008iros}. These local feature descriptor-based methods estimate a complete 3D relative pose between the loop closures. However, they have a high computational cost to identify such feature descriptors from 3D point clouds. They are sensitive to the radially increasing sparsity of point clouds obtained from a standard spinning \lidar{}.

Several methods~\citep{magnusson2009icra-aldf,kim2024ral,angelina2018cvpr,roehling2015iros} take an orthogonal approach by computing one global descriptor per \lidar{} scan.~\cite{magnusson2009icra-aldf} superimpose a voxel grid on the point clouds and approximate a normal distribution over points in each grid cell. They compute a global descriptor for the point clouds by computing a histogram over these normal distributions. More recently,~\cite{kim2024ral} proposed a lightweight global descriptor with \solid{} by representing point clouds in polar coordinates and discretizing them into range-elevation and azimuth-elevation directions, respectively, with each discrete bin storing the number of points. Such global descriptors are easy to store in a simple database, resulting in a faster matching process to retrieve loop closure candidates. However, global descriptor-based methods typically require revisits within small error margins around the reference pose to limit the variation in the global descriptor. They often do not generalize when detecting loop closures across \lidars{} with different scanning patterns and resolutions.

\subsection{Two-Dimensional Projection-Based Methods}
Many recent works focus on speeding up the recognition by describing 3D point clouds by a 2D projection~\citep{kim2018iros,li2021icra-ligl,luo2023iccv,yuan2024tro,xu2023tro}. Even though such a projection results in loss of geometric information, these methods perform comparably and sometimes even better than their full 3D counterparts, especially in outdoor scenes. Among the two-dimensional projection-based methods, the cylindrical projection into range images~\citep{steder2010icra,steder2011iros,shan2021icra,chen2020rss} and the orthogonal projection into \bev{} images~\citep{luo2021ral,kim2021tro,lu2022iros,yuan2023icra,wang2020iros} are widely used 2D projections.

Range image projections are equivariant to azimuthal rotations due to the underlying cylindrical projection. They can recover the relative yaw between loop closures but suffer from scale distortions due to lateral shifts in the \lidar{} viewpoint. On the other hand, \bev{} projections preserve 2D geometry along the local \ground{}, which is vital for autonomous ground robots. Among the \bev{} projection methods, the elevation map is widely used~\citep{kim2018iros,kim2021tro,yuan2023icra,luo2023iccv}. An elevation map preserves the maximum elevation within each discrete pillar in the \bev{} representation.

\scancontext{} by~\cite{kim2018iros} is a popular \lidar{}-based loop closure approach. It computes a global descriptor for each \lidar{} scan using the elevation map in polar coordinates. The polar coordinates make the descriptor equivariant to in-plane rotations, thereby allowing the loop closure alignment module to estimate the relative yaw between \lidar{} scans from the same location. However, the polar coordinate representation can be sensitive to in-plane translations. \scancontext{}++~\citep{kim2021tro} improves upon this drawback of \scancontext{} and computes elevation maps in Cartesian coordinates to augment the polar elevation maps. However, preserving the maximum elevation of the points after the \bev{} projection makes the two approaches sensitive to the vertical resolution and \fov{} of the \lidar{}\@.

In contrast, BVMatch~\citep{luo2021ral} proposes a density-preserving \bev{} projection of point clouds and a Log Gabor filtering step. It requires training a bag-of-words model to retrieve loop closure candidates, making it less suitable for real-time applications such as SLAM\@. Our approach also uses a density-preserving \bev{} projection. It generates a database of local feature descriptors from the \bev{} image online using the HBST data structure~\citep{schlegel2018ral} for efficient operation. The use of local feature descriptors for loop closure detection has also been previously explored in the 2D \lidar{} domain with traditional occupancy grid maps~\citep{blanco2013robotica}.

\cite{yuan2023icra} propose utilizing a set of geometric primitives,~\ie{}, triangles with unique side lengths, to describe a point cloud. They obtain the feature vertices of such triangles by a local \bev{} projection of points within voxels that lie at the boundaries of large planar regions in the environment. BTC~\citep{yuan2024tro} improves upon this work by proposing a binary descriptor for a detailed representation of local geometry. They combine the triangle descriptors and the newly proposed binary descriptors to perform loop closure retrieval. The works by~\cite{yuan2023icra,yuan2024tro} use a \map{} representation of the environment by accumulating a fixed number of consecutive scans. This helps them tackle the sparsity issue present with rotating 3D \lidar{}. The \maps{} also make their approach generalizable to \lidar{} sensors with non-repetitive scanning patterns and small \fov{}s. We also use a \map{} representation to detect loop closures. However, unlike STD and BTC, we accumulate scans until a minimum displacement of the platform to ensure that the \maps{} capture a sufficient portion of the scene.

\begin{figure*}[t]
	\fontsize{9}{9}\selectfont
	\def\svgwidth{0.99\linewidth}
	\captionsetup{type=figure}
	\input{figure_2.pdf_tex}
	\captionof{figure}{Overview of our pipeline for loop closure detection and alignment. Given a \map{}~$\pts{\mM}{m}{}$ generated using a travel displacement criterion: (1) We estimate a ground alignment transform~$\T{g}{m}$ that aligns the \ground{} with the \xy{} of the reference frame. (2) We project the ground-aligned \map{}~$\pts{\mM}{m}{\,g}$ onto the local \xy{} to obtain a 2D density image~$\mI_m$. (3) From this density image, we extract ORB feature descriptors~$\mD_m$ and apply self-similarity pruning to remove redundant features. (4) We incrementally insert the pruned feature descriptors into a binary search tree database, which returns feature matches between the query \map{}~$\pts{\mM}{m}{\,g}$ with feature descriptors~$\mD_m$ and reference \maps{}~$\{\pts{\mM}{r}{\,g}\}$ with feature descriptors~$\{\mD_r\}$. These matches form the initial loop closure candidates. (5) We verify candidates geometrically using RANSAC, obtaining the number of inliers and the corresponding 2D transform~$\mq{T}_{\text{\bev{}}}$. We classify candidates as a loop closures if the number of inliers exceeds a threshold. (6) Finally, our pipeline outputs a relative 3D transform~$\T{m}{r}$ that provides an initial alignment of the two \maps{}, which a subsequent \pg{} optimizer can refine.}\label{fig:2}
	\vspace{-5pt}
\end{figure*}%

\subsection{Learning-Based Approaches}
The recent popularity of deep neural networks, attributed to the widespread availability of specialized hardware and software to train such networks, has generated much interest from a \lidar{}-based loop closing perspective~\citep{dube2018rss,chen2020rss,ma2024tii,komorowski2021wacv,ma2022ral}. PointNet~\citep{qi2017cvpr} and PointNet++~\citep{qi2017nips} use multi-layer perceptron networks to detect local feature descriptors from point clouds. PointNetVLAD~\citep{angelina2018cvpr} proposes an end-to-end trained neural network that combines the local feature descriptors obtained from PointNet into a global descriptor using NetVLAD~\citep{arandjelovic2016cvpr-ncaf}. SegMap~\citep{dube2018rss} segments point clouds using a region-growing algorithm and computes data-driven descriptors of these segments to detect loop closures.

\bev{} projections are also popular with learning-based approaches~\citep{kim2019ral,luo2023iccv,luo2025tro,xu2021ral}.~\cite{kim2019ral} repurpose the \scancontext{}~\citep{kim2018iros} image as a three-channel image using a jet colormap over a range of structural heights. They train a convolutional neural network using these \scancontext{} images and perform localization as a classification task.~\cite{xu2021ral} propose an encoder-decoder network to extract descriptors from differentiable \scancontext{} images.

\cite{vidanapathirana2022icra} use a global descriptor obtained from a sparse convolutional neural network to perform loop closures. They propose a local-consistency loss to train their approach to obtain consistent local features across revisits, which they later combine into a global descriptor using a pooling and normalization strategy.

BEVPlace~\citep{luo2023iccv} applies group convolutions corresponding to the~$\SO{2}$ rotational group to obtain rotation equivariant local features on \bev{} density images and computes a global descriptor for place retrieval using NetVLAD\@. \bevplace{}~\citep{luo2025tro} improves upon BEVPlace by proposing a novel rotation equivariant module, which passes rotation-warped \bev{} density images through a convolutional neural network to extract the local features.

Recently,~\cite{ramezani2024ral} proposed a place recognition method based on an attentional graph neural network representation. Their approach leverages the topological relationships between consecutive \lidar{} scans within a pose-graph SLAM system by generating sub-graphs according to a travel distance criterion. They use a multi-layer perceptron to encode the odometry poses and global descriptor associated with every scan within each of the sub-graphs. For a candidate pair of sub-graphs, they construct a fully connected multiplex graph from the encoded nodes and process it through an attentional graph neural network, P-GAT, to perform place recognition.

Even though learning-based approaches achieve impressive performance regarding loop closure precision and recall metrics, they typically require a computationally expensive offline pre-training step on a GPU, affecting their application to real-time loop closure detection in \slam{}.

Our method also exploits the spatio-temporal information across consecutive \lidar{} scans by constructing \maps{} as the primary representation for loop closure detection, rather than relying on individual scans. Following our earlier work~\citep{gupta2024icra}, we generate \maps{} based on a travel displacement criteria, in contrast to~\cite{yuan2023icra,yuan2024tro}, who accumulate a fixed number of scans. This strategy helps us overcome the sparsity of rotating \lidar{} and make our method agnostic to the \lidar{} sensor's scanning pattern, \fov{}, and resolution. We use a density-preserving \bev{} projection of the \maps{} on the local \ground{} to reduce dimensionality for computational efficiency. Subsequently, we compute ORB feature descriptors~\citep{rublee2011iccv} from the \bev{} density images and store them in an HBST database~\citep{schlegel2018ral} for place recognition. We mitigate the negative impact of perceptual aliasing on our feature descriptor-matching algorithm by performing a self-similarity pruning of the ORB feature descriptors~\citep{bosse2004ijrr}. Our loop closure pipeline provides a complete 3D global alignment between the \maps{}.

\section{Our Methodology}\label{sec:main}

We propose an approach to detect loop closures between \maps{} using their \bev{} representation that works with both vehicle-mounted and handheld \lidar{} sensor platforms. It also provides a complete 3D initial guess of the relative pose between \maps{} involved in the loop closures. We present an overview of our pipeline in~\figref{fig:2}.

We explain in~\secref{sec:local_maps} the procedure to generate \maps{} from \lidar{} point clouds based on a travel displacement criterion. Then, in~\secref{sec:ground_alignment}, we present an approach to detect and align the \ground{} in the \maps{} to the \xy{} of the local reference frame. This allows us to compute a \bev{} representation even in scenarios with a non-planar motion of the \lidar{} sensor, as explained in~\secref{sec:density_images}. We compute binary descriptors for point features on these \bev{} images, as presented in~\secref{sec:features}. To enable place recognition using these binary feature descriptors, we store them in an HBST database~\citep{schlegel2018ral} for subsequent matching with query \maps{} as presented in~\secref{sec:database}. We obtain a set of candidate loop closures for new feature descriptors from a query \map{} by matching these feature descriptors to the database. We perform loop closure validation and geometric alignment in 2D between the matched feature descriptors using a RANSAC~\citep{fischler1981cacm} scheme described in~\secref{sec:alignment}. This 2D alignment, together with the ground alignment for each \map{}, provides the relative 3D transform aligning each pair of loop-closed \maps{}.

\subsection{Creation of Local Maps}\label{sec:local_maps}

\begin{figure}[t]
	\fontsize{9}{9}\selectfont
	\def\svgwidth{0.99\linewidth}
	\captionsetup{type=figure}
\begingroup%
  \makeatletter%
  \providecommand\color[2][]{%
    \errmessage{(Inkscape) Color is used for the text in Inkscape, but the package 'color.sty' is not loaded}%
    \renewcommand\color[2][]{}%
  }%
  \providecommand\transparent[1]{%
    \errmessage{(Inkscape) Transparency is used (non-zero) for the text in Inkscape, but the package 'transparent.sty' is not loaded}%
    \renewcommand\transparent[1]{}%
  }%
  \providecommand\rotatebox[2]{#2}%
  \newcommand*\fsize{\dimexpr\f@size pt\relax}%
  \newcommand*\lineheight[1]{\fontsize{\fsize}{#1\fsize}\selectfont}%
  \ifx\svgwidth\undefined%
    \setlength{\unitlength}{267.81581849bp}%
    \ifx\svgscale\undefined%
      \relax%
    \else%
      \setlength{\unitlength}{\unitlength * \real{\svgscale}}%
    \fi%
  \else%
    \setlength{\unitlength}{\svgwidth}%
  \fi%
  \global\let\svgwidth\undefined%
  \global\let\svgscale\undefined%
  \makeatother%
  \begin{picture}(1,0.51316409)%
    \lineheight{1}%
    \setlength\tabcolsep{0pt}%
    \put(0,0){\includegraphics[width=\unitlength,page=1]{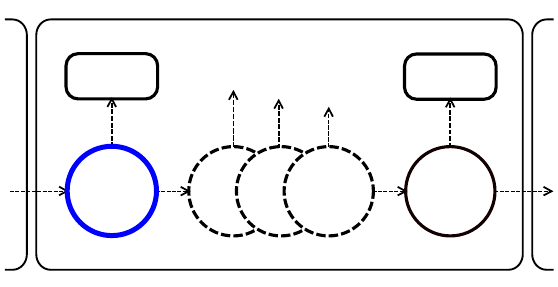}}%
    \put(0.20037917,0.36322107){\makebox(0,0)[t]{\lineheight{1.25}\smash{\begin{tabular}[t]{c}$\pts{\mP}{i}{}$\end{tabular}}}}%
    \put(0,0){\includegraphics[width=\unitlength,page=2]{figure_3.pdf}}%
    \put(0.5009607,0.04578811){\makebox(0,0)[t]{\lineheight{1.25}\smash{\begin{tabular}[t]{c}$\rightarrow$~sequential stream of \lidar{} point clouds~$\rightarrow$\end{tabular}}}}%
    \put(0.80704732,0.36322108){\makebox(0,0)[t]{\lineheight{1.25}\smash{\begin{tabular}[t]{c}$\pts{\mP}{i+k-1}{}$\end{tabular}}}}%
    \put(0.80704731,0.15772339){\makebox(0,0)[t]{\lineheight{1.25}\smash{\begin{tabular}[t]{c}$\T{w}{i+k-1}$\end{tabular}}}}%
    \put(0.20037919,0.15831002){\makebox(0,0)[t]{\lineheight{1.25}\smash{\begin{tabular}[t]{c}$\T{w}{i}$\end{tabular}}}}%
    \put(0.50139827,0.4937871){\makebox(0,0)[t]{\lineheight{1.25}\smash{\begin{tabular}[t]{c}Local Map~($\pts{\mM}{m}{}$)\end{tabular}}}}%
  \end{picture}%
\endgroup%

	\captionof{figure}{A block diagram showcasing the composition of a \map{}~$\pts{\mM}{m}{}$, generated using point clouds \mbox{$\{\pts{\mP}{i}{}, \ldots, \pts{\mP}{i+k-1}{}\}$} registered with corresponding odometry pose estimates $\{\T{w}{i}, \ldots, \T{w}{i+k-1}\}$. The \map{} is referenced to the local coordinate frame of the first scan in this \map{}, as highlighted in blue.}\label{fig:3}
	\vspace{-5pt}
\end{figure}

\begin{figure*}[t]
	\begin{center}
		\def\svgwidth{0.95\linewidth}
		\captionsetup{type=figure}
\begingroup%
  \makeatletter%
  \providecommand\color[2][]{%
    \errmessage{(Inkscape) Color is used for the text in Inkscape, but the package 'color.sty' is not loaded}%
    \renewcommand\color[2][]{}%
  }%
  \providecommand\transparent[1]{%
    \errmessage{(Inkscape) Transparency is used (non-zero) for the text in Inkscape, but the package 'transparent.sty' is not loaded}%
    \renewcommand\transparent[1]{}%
  }%
  \providecommand\rotatebox[2]{#2}%
  \newcommand*\fsize{\dimexpr\f@size pt\relax}%
  \newcommand*\lineheight[1]{\fontsize{\fsize}{#1\fsize}\selectfont}%
  \ifx\svgwidth\undefined%
    \setlength{\unitlength}{579.30648684bp}%
    \ifx\svgscale\undefined%
      \relax%
    \else%
      \setlength{\unitlength}{\unitlength * \real{\svgscale}}%
    \fi%
  \else%
    \setlength{\unitlength}{\svgwidth}%
  \fi%
  \global\let\svgwidth\undefined%
  \global\let\svgscale\undefined%
  \makeatother%
  \begin{picture}(1,0.57720598)%
    \lineheight{1}%
    \setlength\tabcolsep{0pt}%
    \put(0,0){\includegraphics[width=\unitlength,page=1]{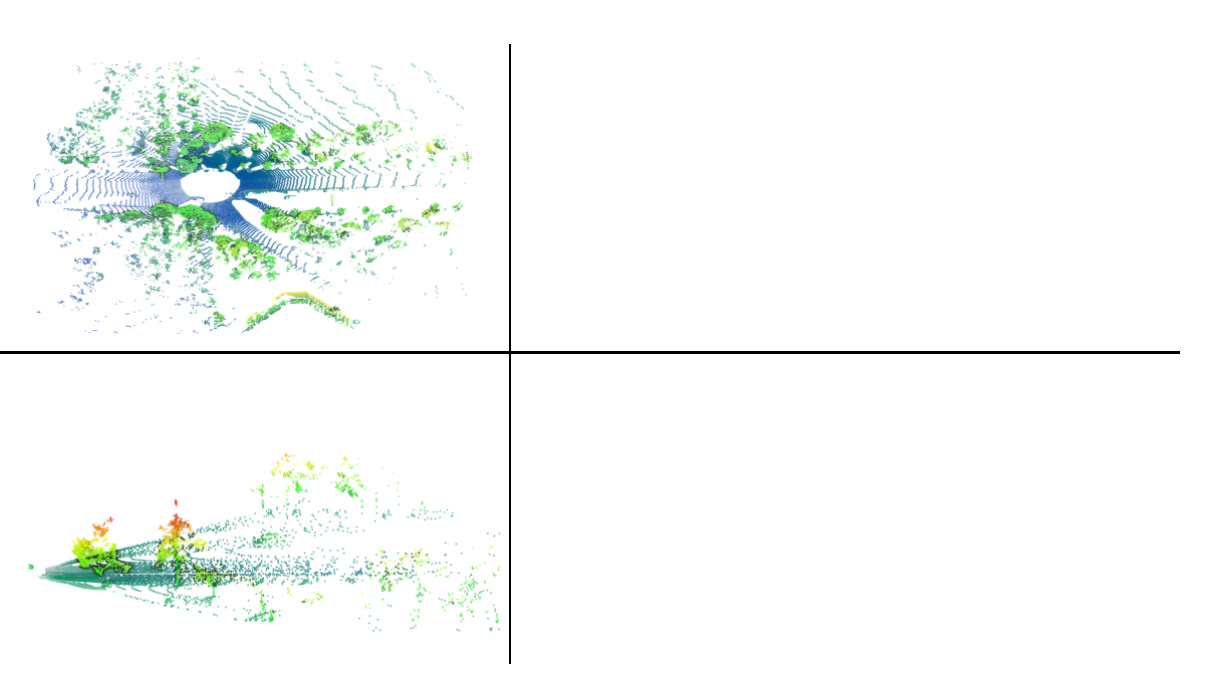}}%
    \put(0.01287778,0.429317){\rotatebox{90}{\makebox(0,0)[t]{\lineheight{1.25}\smash{\begin{tabular}[t]{c}Ouster OS2-128\end{tabular}}}}}%
    \put(0.01303786,0.13634242){\rotatebox{90}{\makebox(0,0)[t]{\lineheight{1.25}\smash{\begin{tabular}[t]{c}Livox Avia\end{tabular}}}}}%
    \put(0,0){\includegraphics[width=\unitlength,page=2]{figure_4.pdf}}%
    \put(0.29552769,0.26517891){\color[rgb]{0.60392157,0,0}\makebox(0,0)[t]{\lineheight{1.25}\smash{\begin{tabular}[t]{c}\textbf{hard to align}\end{tabular}}}}%
    \put(0.21777932,0.56522031){\makebox(0,0)[t]{\lineheight{1.25}\smash{\begin{tabular}[t]{c}single scan\end{tabular}}}}%
    \put(0.72041712,0.56522031){\makebox(0,0)[t]{\lineheight{1.25}\smash{\begin{tabular}[t]{c}\map{}\end{tabular}}}}%
    \put(0.69963861,0.26517891){\color[rgb]{0.60392157,0,0}\makebox(0,0)[t]{\lineheight{1.25}\smash{\begin{tabular}[t]{c}\textbf{easier to align}\end{tabular}}}}%
  \end{picture}%
\endgroup%

		\captionof{figure}{A comparison of data from \lidars{} with different scanning patterns and \fov{}s. The first row shows a single scan from an \ouster{} \lidar{} with a~360$\degrees~\times$~22.5$\degrees$~\fov{} and its corresponding \map{}. The second row shows a \livox{} \lidar{} with a~70$\degrees~\times$~77$\degrees$~\fov{} and its corresponding \map{}. The similar structural features across \maps{} are highlighted with ellipses.}\label{fig:4}
	\end{center}
	\vspace{-10pt}
\end{figure*}

Our approach uses local point cloud maps of the environment to perform loop closures. By aggregating downsampled \lidar{} scans over time, we exploit the local consistency of sequential odometry estimates to generate these \maps{}. We use \kiss{}~\citep{vizzo2023ral} to obtain \lidar{} odometry for our pipeline.

Given an online stream of sequential point clouds \mbox{$\{\pts{\mP}{1}{}, \ldots, \pts{\mP}{n}{}\}$} in the \lidar{} frame and their corresponding 3D pose estimates \mbox{$\{\T{w}{1}, \ldots, \T{w}{n}\}$} in the odometry frame~$w$, we first transform the point clouds into the odometry frame, resulting in point clouds~\mbox{$\{\pts{\mP}{1}{\,w}, \ldots, \pts{\mP}{n}{\,w}\}$}.

Starting from a point cloud with index~$i$, we consider~$k$ subsequent scans until the displacement~\mbox{$\norm{{{}^w}\v{t}_{i+k-1} - {{}^w}\v{t}_i}_2$} exceeds a threshold~$\tau_{c}$, where~\mbox{${{}^w}\v{t}_i \in \RR^3$} is the translational component of the pose estimate~\mbox{$\T{w}{i} \in \SE{3}$}. Once we reach the displacement threshold~$\tau_{c}$, we aggregate this subset of scans~\mbox{$\{\pts{\mP}{i}{\,w}, \ldots, \pts{\mP}{i+k-1}{\,w}\}$} into a \map{}~\mbox{$\pts{\mM}{m}{\,w}$} where~$m$ is the \map{} index. Specifically, we use a voxel grid-based downsampling strategy to generate the \map{}, using a resolution of~$\nu_{\text{map}}$ meters per voxel. Such a downsampling strategy also ensures a uniform spatial distribution of scanned points by retaining at most 20 points per voxel. We repeat this process of \map{} generation sequentially, thereby generating mutually exclusive sets of consecutive point clouds, as shown in~\figref{fig:3}.

We transform each \map{}~\mbox{$\pts{\mM}{m}{\,w}$} into the local reference frame~\mbox{$\T{w}{i}$} of the first scan in the sub-sequence as follows:
\begin{equation}
	\pts{\mM}{m}{} = \T{w}{i}\inv\,\pts{\mM}{m}{\,w}.
\end{equation}

Aggregating consecutive \lidar{} scans into \maps{} provides richer structural information than individual scans, as illustrated in~\figref{fig:4}. This larger spatial context is particularly beneficial in multi-session scenarios, where local environmental changes may otherwise hinder place recognition. Furthermore, \maps{} generated at the same location exhibit higher similarity across different \lidar{} resolutions and scanning patterns, whereas inter-\lidar{} place recognition with individual scans remains a challenging data association problem, even for humans, as illustrated in~\figref{fig:4}. As a result, the use of \maps{} improves robustness for multi-session and cross-platform loop closure detection.

\subsection{Ground Plane Detection and Alignment}\label{sec:ground_alignment}
\begin{figure*}[t]
	\begin{center}
		\def\svgwidth{0.95\linewidth}
		\captionsetup{type=figure}
\begingroup%
  \makeatletter%
  \providecommand\color[2][]{%
    \errmessage{(Inkscape) Color is used for the text in Inkscape, but the package 'color.sty' is not loaded}%
    \renewcommand\color[2][]{}%
  }%
  \providecommand\transparent[1]{%
    \errmessage{(Inkscape) Transparency is used (non-zero) for the text in Inkscape, but the package 'transparent.sty' is not loaded}%
    \renewcommand\transparent[1]{}%
  }%
  \providecommand\rotatebox[2]{#2}%
  \newcommand*\fsize{\dimexpr\f@size pt\relax}%
  \newcommand*\lineheight[1]{\fontsize{\fsize}{#1\fsize}\selectfont}%
  \ifx\svgwidth\undefined%
    \setlength{\unitlength}{587.38714744bp}%
    \ifx\svgscale\undefined%
      \relax%
    \else%
      \setlength{\unitlength}{\unitlength * \real{\svgscale}}%
    \fi%
  \else%
    \setlength{\unitlength}{\svgwidth}%
  \fi%
  \global\let\svgwidth\undefined%
  \global\let\svgscale\undefined%
  \makeatother%
  \begin{picture}(1,0.3184669)%
    \lineheight{1}%
    \setlength\tabcolsep{0pt}%
    \put(0,0){\includegraphics[width=\unitlength,page=1]{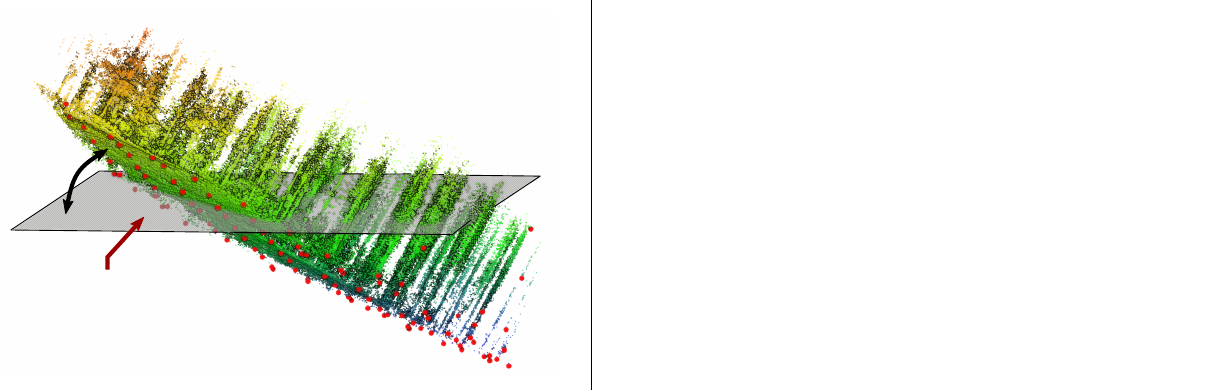}}%
    \put(0.08817931,0.07904271){\makebox(0,0)[t]{\lineheight{1.25}\smash{\begin{tabular}[t]{c}local \xy{}\end{tabular}}}}%
    \put(0.03267818,0.17880804){\color[rgb]{0,0,0}\makebox(0,0)[t]{\lineheight{1.25}\smash{\begin{tabular}[t]{c}$\T{g}{m}$\end{tabular}}}}%
    \put(0.22649321,0.30107331){\makebox(0,0)[t]{\lineheight{1.25}\smash{\begin{tabular}[t]{c}before alignment~($\pts{\mM}{m}{}$)\end{tabular}}}}%
    \put(0,0){\includegraphics[width=\unitlength,page=2]{figure_5.pdf}}%
    \put(0.87045866,0.04903191){\makebox(0,0)[t]{\lineheight{1.25}\smash{\begin{tabular}[t]{c}local \xy{}\end{tabular}}}}%
    \put(0.7566917,0.30107331){\makebox(0,0)[t]{\lineheight{1.25}\smash{\begin{tabular}[t]{c}after alignment~($\pts{\mM}{m}{\,g}$)\end{tabular}}}}%
  \end{picture}%
\endgroup%

		\captionof{figure}{Effect of ground alignment on a \map{} generated from a handheld \lidar{} sensor, recorded in a forest, with non-planar motion. The colors of the points represent their z-coordinates in the local reference frame. \textit{Left}: Our ground alignment approach samples ground points~$\pts{\mG}{m}{}$~(highlighted in red) from the \map{}~$\pts{\mM}{m}{}$ and computes an alignment~$\T{g}{m}$ to the \xy{}. \textit{Right}: The ground-aligned \map{}~$\pts{\mM}{m}{\,g}$.}\label{fig:5}
	\end{center}
	\vspace{-20pt}
\end{figure*}

After generating the point cloud \map{}~$\pts{\mM}{m}{}$, we perform a \bev{} projection. To ensure consistency across multiple revisits, we must project the point cloud \emph{onto a consistent physical plane} in the environment. The \ground{} provides a natural choice for this plane and enables the computation of a stable basis for the \bev{} projection. In our earlier work~\citep{gupta2024icra}, we assumed planar sensor motion so that the \ground{} remained parallel to the \xy{} of the \map{}'s reference frame. However, when the \lidar{} follows a non-planar trajectory, the local \ground{} tilts relative to the \xy{}, as shown in~\figref{fig:5}. To address this, we explicitly identify the \ground{} and align it with the \xy{} of the reference frame.

The first step identifies candidate ground points in the \map{}. Semantic segmentation networks can label ground in point clouds~\citep{xu2021iccv-rada,paigwar2020iros}, but they require offline GPU training and add significant computational costs. Real-time CPU-based methods such as Patchwork~\citep{lim2021ral-patchwork} and Patchwork++~\citep{lee2022iros} operate on individual scans, so they require per-scan segmentation and careful parameter tuning. In contrast, our \bev{} projection requires only one approximate estimate of the local \ground{} to align the entire \map{} at once, not a precise segmentation for every scan.

We adopt a simple sampling strategy: in a 3D \lidar{} \map{}, the lowest-elevation points usually correspond to the ground~\citep{hu2014ivs}. This assumption generally holds for typical outdoor sensing platforms with known extrinsics, where the \ground{} lies within the \lidar{}'s field of view. It also remains valid when the \map{}'s \xy{} is not perfectly aligned with the \ground{}, since a \lidar{} cannot scan below the ground surface.

We divide the \map{}~$\pts{\mM}{m}{}$ into a 2D grid over the \xy{} with~1.0\,m cells. In each cell, we retain only the point with the minimum z-coordinate, yielding a sparse set of candidate ground points, as shown in~\figref{fig:5}. This set may still contain non-ground points where the ground is occluded or where the \xy{} is misaligned with the \ground{}. To filter these, we run principal component analysis~(PCA) on the normal vectors associated with the sampled points and retain only samples whose normals have a cosine similarity larger than $0.95$ with the first principal component. This step removes most non-ground points and yields the ground sample set~\mbox{$\pts{\mG}{m}{}$}. The first principal component also provides a strong initial estimate of the ground orientation.

We convert this PCA estimate into an initial transform~$\mq{T}_{\text{init}} \in \SE{3}$ for ground alignment. Specifically, we compute an axis–angle rotation that aligns the first principal component with the z-axis of the \map{} frame:
\begin{align}
	&\v{a} = \v{v_{\text{pc}}} \times {\dvectort{0}{0}{1}^{\tr}},\\
	&\theta = \arccos{\big(\v{v_{\text{pc}}} \cdot \dvectort{0}{0}{1}^{\tr}\big)},
\end{align}
where,~$\v{v_{\text{pc}}}$ denotes the first principal component,~$\v{a}$ the axis of rotation, and~$\theta$ the rotation angle. We convert this axis-angle pair into a rotation matrix~\mbox{$\m{R}_{\text{init}} \in \SO{3}$} using Rodrigues' formula. We also compute a mean z-shift:
\begin{equation}
	z_{\text{init}} = \v{r}_{2} \cdot \bar{\v{g}}_m,
\end{equation}
where,~$\v{r}_{2}$ is the last row of~$\m{R}_{\text{init}}$ and~$\bar{\v{g}}_m$ is the mean of~$\pts{\mG}{m}{}$. We combine~$\m{R}_{\text{init}}$ and~$z_{\text{init}}$ into the initial transform~$\mq{T}_{\text{init}}$.

To make the ground alignment robust to outliers, we refine $\mq{T}_{\text{init}}$ with a least-squares optimization:
\begin{equation}
	\T{g}{m} = \argmin_{\mq{T}_{\text{init}}} \sum_{\forall \bg_k \in \pts{\mG}{m}{}} {w_k \big({\hat{\v{n}}_{xy}} \cdot (\mq{T}_{\text{init}}\,\bg_k) \big)}^{2},
	\label{eq:least-sq}
\end{equation}
where,~\mbox{$\mq{T}_{\text{init}}\,\bg_k = \dvectort{x_k^{\prime}}{y_k^{\prime}}{z_k^{\prime}}^{\tr}$} denotes the~$\SE{3}$ group action on~$\RR^3$,~$\hat{\v{n}}_{xy} = \dvectort{0}{0}{1}^{\tr}$ is the normal vector of the \xy{}, and~$w_k$ is a Gaussian weight that reduces the influence of spurious outliers. We iterate this optimization, stopping after ten iterations or once the pose correction between iterations falls below a threshold.

We compute the Jacobian~$\mJ$ of the residual term \mbox{${\hat{\v{n}}_{xy}} \cdot (\mq{T}_{\text{init}}\,\bg_k)$} in~\eqref{eq:least-sq} with respect to a perturbation~\mbox{$\Delta \v{x} = \zvectort{\Delta \v{t}^{\tr}}{\Delta \v{\omega}^{\tr}}^{\tr}$} around~$\mq{T}_{\text{init}}$ on the~$\SE{3}$ manifold:
\begin{equation}
	\mJ = \svectort{\,0\,}{0\,}{1\,}{x_k^{\prime}\,}{-y_k^{\prime}\,}{0\,}.
	\label{eq:jacobian}
\end{equation}

The Jacobian indicates that the optimization only updates the z-axis translation and the roll~(x-axis) and pitch~(y-axis), which directly correct the ground alignment.

Finally, we apply the ground-aligning transform~$\T{g}{m}$ to the \map{}~$\pts{\mM}{m}{}$ to obtain the aligned \map{}~$\pts{\mM}{m}{\,g}$ whose \ground{} lies on the \xy{}. The alignment step increases robustness to non-planar \lidar{} motion profiles, particularly for handheld platforms. It establishes the \ground{} as a robust reference plane for \bev{} projections during long-term revisits, since the ground surface typically remains consistent over time. We show an example of this alignment in~\figref{fig:5} and provide a detailed analysis in~\secref{sec:alignment_study}.

\subsection{Density-Preserving Bird's-Eye-View Projection}\label{sec:density_images}
After establishing the primary representation of the environment in~\secref{sec:local_maps} and~\secref{sec:ground_alignment}, the next step is to extract informative yet distinct local feature descriptors to perform place recognition for loop closures. Rather than computing features directly on the 3D point clouds of \maps{}, which is computationally expensive given the spatial extent of such point clouds, we use a 2D \bev{} projection of the \maps{} as an intermediate representation for loop closure detection.

We perform the \bev{} projection by simply dropping the z-coordinate of all the individual points in the ground-aligned \maps{}~\mbox{$\pts{\mM}{m}{\,g}$}. Furthermore, we discretize the~$\RR^2$ projection space with a resolution of~$\nu_{\text{b}}$ meters. This results in a 2D Cartesian grid~$\mN_m(u, v)$ of size $W_m \times H_m$ where,
\begin{align}
	W_m & = \left\lfloor\frac{x_m^u}{\nu_{\text{b}}}\right\rfloor - \left\lfloor\frac{x_m^l}{\nu_{\text{b}}}\right\rfloor + 1, \\
	H_m & = \left\lfloor\frac{y_m^u}{\nu_{\text{b}}}\right\rfloor - \left\lfloor\frac{y_m^l}{\nu_{\text{b}}}\right\rfloor + 1,
\end{align}
\begin{equation}
	\zvector{x_m^u}{y_m^u} = \max_{x, y}\pts{\mM}{m}{\,g}\ ;\ \zvector{x_m^l}{y_m^l} = \min_{x, y}\pts{\mM}{m}{\,g}.
\end{equation}

However, the dimensionality reduction comes at the cost of losing the complete 3D information about the scene. Many traditional \bev{} projection approaches store the maximum elevation of the points in each cell~\citep{kim2021tro,kim2018iros,li2021icra-ligl,luo2022tiv}. The maximum elevation, however, is sensitive to the distance between the scanner and the surface being scanned, as well as the \lidar{} sensor's \fov{}. In our pipeline, we instead store the \emph{point density} in each discrete 2D cell after the projection, which is less sensitive to viewpoint changes~\citep{luo2021ral}.

Therefore, each cell in this grid~\mbox{$\mN_m(u, v) \in \NN_0$} stores the count of projected points within that cell. The \bev{} image~$\mI_m(u, v)$ of the \map{}~$\pts{\mM}{m}{\,g}$ is then defined as the relative point density in each cell of the grid as follows:
\begin{equation}
	\mI_m(u, v) = \frac{\mN_m(u, v) - N_{\text{min}}}{N_{\text{max}} - N_{\text{min}}} \in \RR^{W_m \times H_m},
\end{equation}
\begin{equation}
	N_{\text{max}} = \max{\mN_m(u, v)}; \ N_{\text{min}} = \min{\mN_m(u, v)}.
\end{equation}

To mitigate the undesired influence of dynamic objects that accumulate during \map{} generation, we explicitly set all image pixels~$\mN_m(u, v)$ with a relative density lower than 5\% to zero. This choice reflects the fact that most dynamic objects commonly found in urban environments, such as vehicles, pedestrians, and cyclists, have small vertical footprints and therefore produce low-density values.

\subsection{Feature Detection and Pruning Strategy}\label{sec:features}
The \bev{} projection reduces the dimensionality of the \map{}, making feature detection for place recognition computationally efficient. The image-like \bev{} density representation enables us to apply well-established computer vision techniques to extract distinctive feature descriptors. Since these \bev{} density images preserve the geometrical, floorplan-like structure of the environment, we use ORB~\citep{rublee2011iccv} feature descriptors to capture relevant features. Unlike camera images, \bev{} images generated via orthographic projection have no scale ambiguity. We take advantage of this property by computing the ORB feature descriptors~$\mD_m$ without applying scale-invariance adjustments, further improving computational efficiency.

\begin{figure}[t]
	\def\svgwidth{0.99\linewidth}
	\captionsetup{type=figure}
\begingroup%
  \makeatletter%
  \providecommand\color[2][]{%
    \errmessage{(Inkscape) Color is used for the text in Inkscape, but the package 'color.sty' is not loaded}%
    \renewcommand\color[2][]{}%
  }%
  \providecommand\transparent[1]{%
    \errmessage{(Inkscape) Transparency is used (non-zero) for the text in Inkscape, but the package 'transparent.sty' is not loaded}%
    \renewcommand\transparent[1]{}%
  }%
  \providecommand\rotatebox[2]{#2}%
  \newcommand*\fsize{\dimexpr\f@size pt\relax}%
  \newcommand*\lineheight[1]{\fontsize{\fsize}{#1\fsize}\selectfont}%
  \ifx\svgwidth\undefined%
    \setlength{\unitlength}{256.56439113bp}%
    \ifx\svgscale\undefined%
      \relax%
    \else%
      \setlength{\unitlength}{\unitlength * \real{\svgscale}}%
    \fi%
  \else%
    \setlength{\unitlength}{\svgwidth}%
  \fi%
  \global\let\svgwidth\undefined%
  \global\let\svgscale\undefined%
  \makeatother%
  \begin{picture}(1,0.52653023)%
    \lineheight{1}%
    \setlength\tabcolsep{0pt}%
    \put(0,0){\includegraphics[width=\unitlength,page=1]{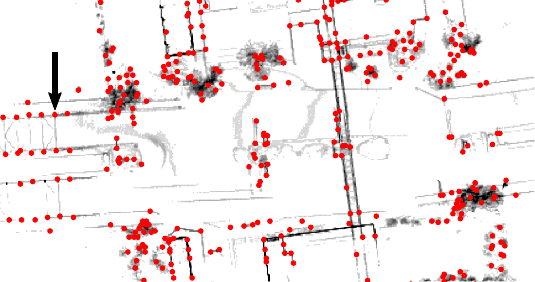}}%
    \put(0.10278045,0.49532968){\makebox(0,0)[t]{\lineheight{1.25}\smash{\begin{tabular}[t]{c}ORB\\features\end{tabular}}}}%
  \end{picture}%
\endgroup%

	\captionof{figure}{A \bev{} density image of a \map{}, where darker pixels indicate higher point density. Red dots mark the ORB features extracted from the image.}\label{fig:6}
	\vspace{-2.5pt}
\end{figure}

\figref{fig:6} shows an example of ORB features extracted from a \bev{} density image. We observe that most ORB features concentrate around high-density regions with strong gradients. These strong responses typically arise from static structures with large vertical footprints, such as building facades, trees, and light poles. In contrast, low-density regions, often corresponding to small bushes, vegetation, or dynamic objects that occasionally pass through the low-density filter, contribute few ORB features to the database. This leads to a robust behavior of our algorithm as it is not adversely affected by dynamic objects in the environment.

ORB feature descriptors remain salient within a local neighborhood of the density image, but not necessarily across the entire image. As a result, environments with repetitive structures can generate self-similar feature descriptors that can confuse the feature-matching stage and cause false positives.

To improve robustness against perceptual aliasing, we prune the ORB feature descriptors within each density image using a self-similarity check inspired by~\cite{bosse2004ijrr}. Specifically, we compute the nearest-neighbor match for each ORB feature descriptor based on the Hamming distance and discard any feature whose closest match lies below a threshold of~\mbox{$\tau_{pr}=\text{35}$\,bits}. This procedure eliminates self-similar feature descriptors within the same density image. This approach resembles Lowe's ratio test~\citep{lowe2004ijcv}, but we apply it intra-image rather than inter-image, which avoids the cost of finding second-best matches across a larger feature descriptor database.

\begin{figure}[t]
	\def\svgwidth{0.99\linewidth}
	\captionsetup{type=figure}
\begingroup%
  \makeatletter%
  \providecommand\color[2][]{%
    \errmessage{(Inkscape) Color is used for the text in Inkscape, but the package 'color.sty' is not loaded}%
    \renewcommand\color[2][]{}%
  }%
  \providecommand\transparent[1]{%
    \errmessage{(Inkscape) Transparency is used (non-zero) for the text in Inkscape, but the package 'transparent.sty' is not loaded}%
    \renewcommand\transparent[1]{}%
  }%
  \providecommand\rotatebox[2]{#2}%
  \newcommand*\fsize{\dimexpr\f@size pt\relax}%
  \newcommand*\lineheight[1]{\fontsize{\fsize}{#1\fsize}\selectfont}%
  \ifx\svgwidth\undefined%
    \setlength{\unitlength}{316.9027758bp}%
    \ifx\svgscale\undefined%
      \relax%
    \else%
      \setlength{\unitlength}{\unitlength * \real{\svgscale}}%
    \fi%
  \else%
    \setlength{\unitlength}{\svgwidth}%
  \fi%
  \global\let\svgwidth\undefined%
  \global\let\svgscale\undefined%
  \makeatother%
  \begin{picture}(1,0.32613836)%
    \lineheight{1}%
    \setlength\tabcolsep{0pt}%
    \put(0,0){\includegraphics[width=\unitlength,page=1]{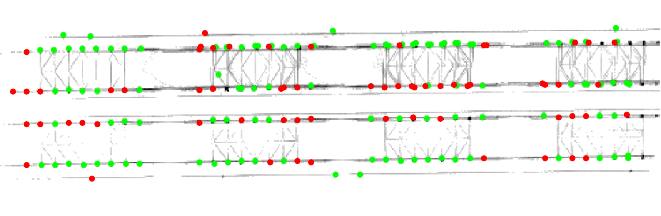}}%
    \put(0.63113185,0.01965485){\makebox(0,0)[lt]{\lineheight{1.25}\smash{\begin{tabular}[t]{l}pruned features\end{tabular}}}}%
    \put(0.19910507,0.01649623){\makebox(0,0)[lt]{\lineheight{1.25}\smash{\begin{tabular}[t]{l}retained features\end{tabular}}}}%
    \put(0,0){\includegraphics[width=\unitlength,page=2]{figure_7.pdf}}%
  \end{picture}%
\endgroup%

	\captionof{figure}{An example of self-similarity feature pruning applied to ORB feature descriptors computed on a \bev{} density image of a bridge with repetitive mechanical structures. Our algorithm prunes the features shown in red and retains those in green.}\label{fig:7}
	\vspace{-2.5pt}
\end{figure}

This pruning step filters out features from repetitive structures in the environment, such as repeating trusses on a bridge, as shown in~\figref{fig:7}. This helps reduce false-positive feature matches across density images resulting from perceptual aliasing, thereby lowering the risk of incorrect loop closure detections. We present a detailed analysis of this pruning strategy and its impact on loop closure detection in~\secref{sec:pruning_study}.

\subsection{Feature Database}\label{sec:database}
Once we have the unique features within a \bev{} density image, we create a database to serve as a reference for place recognition. We leverage the binary domain of the ORB descriptors, using the Hamming distance embedding binary search tree~(HBST)~\citep{schlegel2018ral} to store the set of feature descriptors~$\mD_m$ obtained from each density image~$\mI_m$, along with their corresponding map index~$m$.

The depth of the HBST is limited by the number of bits in the binary descriptor. As a result, a query descriptor will require at most 256 bitwise comparisons with the tree's nodes before reaching a leaf node. Each leaf node can hold a maximum of 100 descriptors, ensuring efficient feature matching. These design choices constrain the computational time for feature matching without imposing practical limitations on the usability of our approach. Unlike a clustering-based bag of words or a neural representation-based database, we use the HBST as an incrementally updated database, requiring no offline pre-training.

\begin{figure}[t]
	\def\svgwidth{0.99\linewidth}
	\captionsetup{type=figure}
\begingroup%
  \makeatletter%
  \providecommand\color[2][]{%
    \errmessage{(Inkscape) Color is used for the text in Inkscape, but the package 'color.sty' is not loaded}%
    \renewcommand\color[2][]{}%
  }%
  \providecommand\transparent[1]{%
    \errmessage{(Inkscape) Transparency is used (non-zero) for the text in Inkscape, but the package 'transparent.sty' is not loaded}%
    \renewcommand\transparent[1]{}%
  }%
  \providecommand\rotatebox[2]{#2}%
  \newcommand*\fsize{\dimexpr\f@size pt\relax}%
  \newcommand*\lineheight[1]{\fontsize{\fsize}{#1\fsize}\selectfont}%
  \ifx\svgwidth\undefined%
    \setlength{\unitlength}{320.00851128bp}%
    \ifx\svgscale\undefined%
      \relax%
    \else%
      \setlength{\unitlength}{\unitlength * \real{\svgscale}}%
    \fi%
  \else%
    \setlength{\unitlength}{\svgwidth}%
  \fi%
  \global\let\svgwidth\undefined%
  \global\let\svgscale\undefined%
  \makeatother%
  \begin{picture}(1,0.66216128)%
    \lineheight{1}%
    \setlength\tabcolsep{0pt}%
    \put(0.6421811,0.28907292){\makebox(0,0)[lt]{\lineheight{1.25}\smash{\begin{tabular}[t]{l}Query \end{tabular}}}}%
    \put(0,0){\includegraphics[width=\unitlength,page=1]{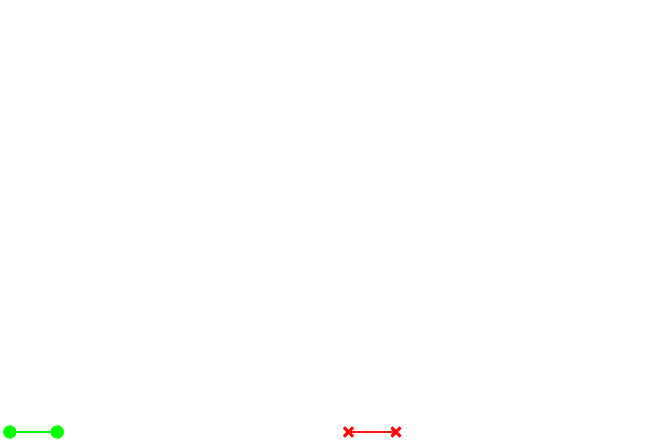}}%
    \put(0.29924765,0.00433942){\makebox(0,0)[t]{\lineheight{1.25}\smash{\begin{tabular}[t]{c}inlier correspondences\end{tabular}}}}%
    \put(0.81206876,0.00433942){\makebox(0,0)[t]{\lineheight{1.25}\smash{\begin{tabular}[t]{c}outlier correspondences\end{tabular}}}}%
    \put(0,0){\includegraphics[width=\unitlength,page=2]{figure_8.pdf}}%
    \put(0.22695444,0.64046391){\makebox(0,0)[t]{\lineheight{1.25}\smash{\begin{tabular}[t]{c}reference density image~($\mI_r$)\end{tabular}}}}%
    \put(0.79625357,0.64046391){\makebox(0,0)[t]{\lineheight{1.25}\smash{\begin{tabular}[t]{c}query density image~($\mI_m$)\end{tabular}}}}%
  \end{picture}%
\endgroup%

	\captionof{figure}{ORB descriptor matches between features from a reference and a query \bev{} density image obtained from the HBST database. The inlier correspondences obtained from the RANSAC-based geometric verification are shown with green lines, and the outlier correspondences with red lines.}\label{fig:8}
\end{figure}

After obtaining a set of descriptors~$\mD_m$ from the query \map{}'s \bev{} density image~$\mI_m$, we search the HBST database to find the closest match for each descriptor~\mbox{$\v{d}_{j,\,m} \in \mD_m$}. Matches are determined using a Hamming distance threshold of 50\,bits between ORB descriptors. Since the binary tree links each stored descriptor to an index~$m$ corresponding to its \map{}, we compile a list of feature matches between the query \map{} and the reference \maps{} in the database. These matches are subsequently verified geometrically.~\figref{fig:8} visualizes the corresponding set of feature matches between two density images, as obtained from the HBST database.

\subsection{Loop Detection and Map Alignment}\label{sec:alignment}
The geometric validation step entails performing a 2D alignment of the matched ORB features. This involves computing a 2D rigid-body transformation that optimally aligns the matched features from the binary search tree using a distance metric. Unlike the general image-alignment problem, this process is constrained to an~$\SE{2}$ transformation rather than a homography. To handle outlier associations caused by the limitations of binary tree-based matching, we employ a RANSAC-based alignment strategy.

We design a RANSAC-based approach that randomly selects two feature pairs from the set of matches between a query~$(\mI_m)$ and a reference~$(\mI_r)$ density image. We use the Kabsch-Umeyama algorithm~\citep{kabsch1976acs,umeyama1991pami} to compute the relative 2D alignment between the feature pairs. We use this alignment to compute the point-wise Euclidean distance between each pair of feature matches. Feature matches with distances larger than~1.5\,m~(3\,pixels for~$\nu_{\text{b}}=\text{0.5}$\,m) are considered outliers. The RANSAC process runs for a fixed number of iterations~$N_{\text{ransac}}$. We require a minimum number of inliers~$\gamma$ from the RANSAC alignment stage to decide whether two \maps{} belong to the same location.

Finally, after the~$N_{\text{ransac}}$ iterations, we compute a Kabsch-Umeyama 2D alignment over the entire set of inlier correspondences. It provides us with a rotation matrix~\mbox{$\m{R} \in \SO{2}$} and a translation vector~\mbox{$\v{t} \in \RR^2$}, which we represent as a homogenous transformation~\mbox{$\mq{T}_{\text{\bev{}}} \in \SE{3}$}, having translation components only along the \xy{}, and a rotation component only about the z-axis. We scale the translation vector by the voxel size~$(\nu_{\text{b}})$ since the features are computed from the discretized density images.

Although~$\mq{T}_{\text{\bev{}}}$ only aligns the ground-aligned \maps{}, we can recover the complete~$\SE{3}$ pose to align the original \maps{} by composing~$\mq{T}_{\text{\bev{}}}$ with the ground-aligning transforms of the individual \maps{}, as shown in~\eqref{eq:3Destimate}.
\begin{equation}
	\T{m}{r} = \T{g}{m}\inv\,\mq{T}_{\text{\bev{}}}\,\T{g}{r}.
	\label{eq:3Destimate}
\end{equation}

\begin{figure}[t]
	\def\svgwidth{0.99\linewidth}
	\captionsetup{type=figure}
\begingroup%
  \makeatletter%
  \providecommand\color[2][]{%
    \errmessage{(Inkscape) Color is used for the text in Inkscape, but the package 'color.sty' is not loaded}%
    \renewcommand\color[2][]{}%
  }%
  \providecommand\transparent[1]{%
    \errmessage{(Inkscape) Transparency is used (non-zero) for the text in Inkscape, but the package 'transparent.sty' is not loaded}%
    \renewcommand\transparent[1]{}%
  }%
  \providecommand\rotatebox[2]{#2}%
  \newcommand*\fsize{\dimexpr\f@size pt\relax}%
  \newcommand*\lineheight[1]{\fontsize{\fsize}{#1\fsize}\selectfont}%
  \ifx\svgwidth\undefined%
    \setlength{\unitlength}{294.45904829bp}%
    \ifx\svgscale\undefined%
      \relax%
    \else%
      \setlength{\unitlength}{\unitlength * \real{\svgscale}}%
    \fi%
  \else%
    \setlength{\unitlength}{\svgwidth}%
  \fi%
  \global\let\svgwidth\undefined%
  \global\let\svgscale\undefined%
  \makeatother%
  \begin{picture}(1,0.65495028)%
    \lineheight{1}%
    \setlength\tabcolsep{0pt}%
    \put(0,0){\includegraphics[width=\unitlength,page=1]{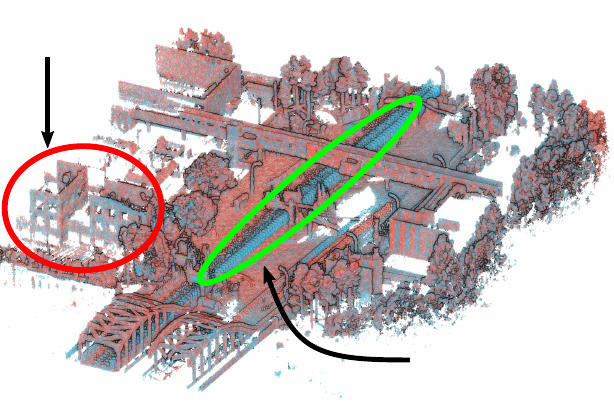}}%
    \put(0.07644952,0.62597558){\makebox(0,0)[t]{\lineheight{1.25}\smash{\begin{tabular}[t]{c}quality of\\alignment\end{tabular}}}}%
    \put(0.81575622,0.07390122){\makebox(0,0)[t]{\lineheight{1.25}\smash{\begin{tabular}[t]{c}trace from a\\dynamic object\end{tabular}}}}%
  \end{picture}%
\endgroup%

	\captionof{figure}{Two \maps{}~(in red and blue) detected as loop closure and aligned using the initial estimate~$\T{m}{r}$ provided by our pipeline. Dynamic objects in the scene, as seen in the area highlighted by a green ellipse, do not affect the alignment quality provided by our approach. The red ellipse highlights the overlap of windowpanes on a building.}~\label{fig:9}
	\vspace{-5pt}
\end{figure}

This 3D alignment serves as an initial estimate for a fine-grained registration of \maps{} during typical \pg{} optimization in a \slam{} pipeline. A qualitative example demonstrating the effectiveness of this initial alignment is illustrated in~\figref{fig:9}. Notably, one of the \maps{} (in blue) in this visualization contains a large trace from a dynamic object (highlighted by the green ellipse), showing that our pipeline can also detect and align loop closures even in the presence of such strong dynamics.

\section{Experimental Setup for Evaluation}\label{sec:exp_setup}

This section presents our experimental setup for evaluating our method and comparing it to existing loop closure detection approaches. We begin by introducing the datasets we use for evaluation and benchmarking. Next, we provide details about the implementation of the baseline methods we include in our comparison. Finally, we describe the quantitative metrics we use to assess the performance of our loop closure detection pipeline and the competing baselines.

\subsection{Datasets}\label{sec:datasets}
We conduct a comprehensive evaluation of our approach on datasets collected using a variety of \lidar{} sensors with different resolutions and scanning patterns. These datasets include multiple sequences recorded in diverse urban environments across various mobile platforms.

\subsubsection{Public Datasets:}
We use the \mulran{}~\citep{kim2020icra} and \helipr{}~\citep{jung2024ijrr} public datasets, both recorded in urban environments using \lidar{} sensors mounted on car rooftops. These datasets are well-suited for intra-session loop closure evaluation due to frequent in-sequence revisits with varying orientations and lane shifts. In addition, both datasets include at least three sequences per environment, making them ideal for evaluating inter-session loop closure detection.

The \mulran{} dataset was recorded using a 64-beam rotating \lidar{}~(Ouster OS1-64) operating at 10\,Hz. Its horizontal \fov{} is limited to~290$\degrees$ due to occlusion from a radar sensor mounted behind it. We use sequences from three environments: KAIST, Riverside, and Sejong.

The \helipr{} dataset features multiple \lidar{} sensors with different \fov{}s, scan patterns, and resolutions mounted on a single vehicle, enabling the evaluation of inter-\lidar{} loop closure detection. From this dataset, we use three of the four available \lidars{}: (i) a 128-beam spinning \lidar{}~(\ouster{}) with~360$\degrees~\times$~22.5$\degrees$~\fov{}, (ii) a hybrid solid-state \lidar{}~(\livox{}) with~70$\degrees~\times$~77$\degrees$~\fov{} and an unusual non-repetitive scanning pattern, and (iii) a solid-state \lidar{}~(\aeva{}) with~120$\degrees~\times$~19.2$\degrees$~\fov{}. We exclude the fourth sensor (Velodyne VLP-16) due to self-occlusion caused by surrounding sensors. The Bridge sequences in this dataset pose an additional challenge due to strong perceptual aliasing from repetitive structures in certain regions.

Moreover, both \mulran{} and \helipr{} datasets include sequences from the KAIST and Riverside environments, recorded approximately 4 years apart. This allows for a realistic evaluation of long-term inter-session loop closure detection, including cross-dataset comparisons between sequences captured with different \lidar{} sensors. The KAIST and Riverside sequences also have some spatial overlap, providing an opportunity to evaluate multi-map alignment in a more challenging setting.

Additionally, we use two sequences from the \nclt{} dataset~\citep{carlevaris-bianco2016ijrr}, recorded using a \velodyne{} \lidar{} mounted on a Segway robot. The motion of the \lidar{} in these sequences is non-planar due to the inverted pendulum-like dynamics of the Segway platform. We select two sequences recorded in the same environment but one year apart to assess the robustness of our loop closure detection approach under viewpoint and temporal variation.

\subsubsection{Self-Recorded Datasets:}
In addition to the public datasets described above, we evaluate our approach on sequences recorded using our own instrumented vehicle and sensor platform. The \car{} sequence was recorded using an Ouster OS1-128 \lidar{} mounted on the rooftop of an instrumented vehicle. We collected this data by driving through a hilly, semi-urban area with forest-lined roads and performing multiple revisits from varying viewpoints.

The \backpack{} sequence was captured using a sensor platform mounted on a backpack and carried through urban streets. This setup uses a Hesai Pandar-128 \lidar{} mounted approximately 2\,m above ground level. Unlike vehicle-mounted systems, the walking motion introduces non-planar sensor trajectories and dynamic changes in the viewpoint. This sequence presents a challenging test case for evaluating our ground alignment strategy under non-planar motion.

The two sequences share a small physical overlap in their environments, which we utilize to evaluate inter-session loop closure detection.

We generate near-ground-truth pose information using the offline \lidar{} bundle adjustment method by~\cite{wiesmann2024arxiv}, with manual inspection of the results to ensure alignment accuracy. This method incorporates RTK-GNSS data for geo-referencing; an information not provided to any loop closure system under evaluation. The process begins with initial pose estimates obtained from \pg{} fusion of \lidar{} odometry~\citep{vizzo2023ral}, aided by either manually added loop closures~(for the \backpack{} sequence) or RTK-GNSS data~(for the \car{} sequence) to enforce global consistency. The final bundle adjustment refines these initial poses by aligning all scans with each other in an offline fashion.

\subsection{Baseline Methods Used for Comparison}\label{sec:baselines}
We compare our approach against seven baseline methods, all of which provide publicly available implementations that we use for evaluation. To ensure a fair comparison, we modify their source code only to return multiple loop closure candidates per query scan rather than just the top-ranked one. Unless stated otherwise, we retain the default parameter settings in each baseline's implementation.

\subsubsection{Our Approach:}
For our approach, we limit the maximum range of each \lidar{} scan to 100\,m. We set the travel displacement threshold~$\tau_{c}$ for generating \maps{} to 100\,m as well. For~\kiss{}~\citep{vizzo2023ral} odometry, we use their latest open-source version~(v1.2.3). We use a voxel size of~\mbox{$\nu_{\text{map}}=\text{0.5}$\,m} for the voxel grid used to generate the \maps{}, and the resolution of the \bev{} density images~\mbox{$\nu_{\text{b}}$} is set to 0.5\,m. We use the default parameters for ORB feature descriptors, except that we turn off the scale invariance since the \bev{} images are orthographic projections. We obtain feature matches from HBST using a Hamming distance threshold of 50\,bits and leave all other HBST parameters at their defaults. Finally, we classify two \maps{} as loop closures if the RANSAC-based alignment yields more than~\mbox{$\gamma=\text{5}$} inliers.

For multi-session experiments, we save the HBST database from the reference session and use it as the database for the query session. We apply the same configuration to evaluate our previous approach, referred to as MapClosure~\citep{gupta2024icra}.

\subsubsection{\scancontext{}~(SC):}
\scancontext{}~\citep{kim2018iros} is a widely used, \sota{} approach for \lidar{}-based loop closure detection. However, it is specifically designed for \lidars{} with a large horizontal \fov{} and a cylindrical scanning pattern. Therefore, we omit evaluation on the \livox{} and \aeva{} sensors from the \helipr{} dataset.

\subsubsection{Stable Triangle Descriptors~(STD) and Binary Triangle Combined Descriptors~(BTC):}
STD by~\cite{yuan2023icra} introduces a triangle descriptor for point cloud \maps{}, which BTC~\citep{yuan2024tro} extends with a binary descriptor for efficient storage and fast retrieval of loop closure candidates. Both methods accumulate 10 consecutive scans into a \map{} to compute descriptors. We use \kiss{} to obtain the pose estimates for constructing these \maps{}. We adopt the default parameter settings from their implementations. For a fair comparison, we also evaluate STD and BTC on the same \maps{} used in our approach, generated with a 100\,m travel displacement criterion. We refer to these variants as STD-100 and BTC-100.

\subsubsection{Spatially Organized and Lightweight Global Descriptor (\solid{}):}
\solid{}~\citep{kim2024ral} is a recent loop closure detection method for 3D \lidar{} sensors that is designed to operate across a wide variety of sensors, irrespective of their scan pattern, \fov{}, or resolution. The method employs a kd-tree-based search strategy to identify candidate loop closures. However, the authors do not provide an implementation of this search procedure in the publicly available codebase. Consequently, we approximate the candidate set by selecting all scans at least hundred frames prior to the current query scan. Following the procedure outlined in their manuscript, we compute the evaluation metrics using cosine distances between the descriptors of the query scan and each candidate.

\subsubsection{LoGG3D-Net:}
LoGG3D-Net~\citep{vidanapathirana2022icra} is a learning-based method that employs a 3D sparse convolutional network to extract consistent local features across different viewpoints. For our evaluation, we use the checkpoint trained on sequences from the \mulran{} dataset, as provided by the authors, and we adopt the default parameter settings from their implementation.

\subsubsection{\bevplace{}:}
\bevplace{} by~\cite{luo2025tro} is a learning-based method that computes a global descriptor for the \bev{} projection of the input scan using a rotation-equivariant module. In our evaluation, we use the checkpoint trained on sequence 00 from the KITTI dataset~\citep{geiger2013ijrr}, as provided by the authors, and we adopt the default parameter settings from their implementation.

\subsection{Reference Loop Closures Between Local Maps}~\label{sec:reference_closures}
Since our method computes loop closures between \maps{}, we also define reference loop closures at the map level for evaluation. In contrast to prior works~\citep{jiang2023icra-ccas,kim2018iros,kim2021tro,yuan2023icra,kim2024ral}, which typically rely on simple distance-based criteria, such heuristics are insufficient in our setting. The distance-based criterion breaks down in scenarios where other objects occlude the previously seen area, the sensor has a limited \fov{}, or the revisits are from a significantly different viewpoint. Additionally, there is inherent ambiguity in selecting the appropriate reference frame for measuring distances between \maps{}. 

To overcome these limitations, we define reference loop closures based on the volumetric overlap of \maps{}, similar to~\cite{gupta2024icra} and~\cite{yuan2024tro}. We generate reference \maps{} using the ground-truth pose information provided by the respective datasets. We ensure that each reference \map{} contains the same set of scans as the corresponding \maps{} produced by our method. We then identify reference loop closures by selecting all pairs of \maps{} that exhibit more than 25\% voxel-level overlap. To avoid trivial short-range closures, we skip the three consecutive \maps{} preceding the query \map{}.

We adopt the same volumetric overlap strategy for cross-sequence evaluation, but reduce the overlap threshold to 10\% to accommodate potential misalignments in the global pose information between sequences.

\subsection{Scan-Level to Map-Level Conversion of Loop Closures for Baselines}~\label{sec:reference_closures_baselines}
The baseline methods in our evaluation operate at different scales; some detect loop closures at the individual scan level~(\scancontext{} and \solid{}), while others work on \maps{} with a fixed number of scans~(STD and BTC). As a result, we cannot directly evaluate them using the reference loop closures defined between \maps{} generated based on a travel displacement criterion. Therefore, we convert their outputs into equivalent map-level loop closures.

For scan-level approaches, we treat a loop closure between scans~$\pts{\mP}{i}{}$ and~$\pts{\mP}{j}{}$ as a loop closure between \maps{}~$\pts{\mM}{k}{}$ and~$\pts{\mM}{l}{}$, where scan~$\pts{\mP}{i}{}$ belongs to map~$\pts{\mM}{k}{}$ and scan~$\pts{\mP}{j}{}$ belongs to map~$\pts{\mM}{l}{}$.

For methods such as STD and BTC, which use smaller \maps{}, we first extract all scan pairs between these \maps{},~$\pts{\mS}{m}{}$ and~$\pts{\mS}{n}{}$, that form a loop closure. We treat each scan pair as an individual scan-level loop closure and convert it into a corresponding map-level loop closure using the same method described above.


\subsection{Evaluation Metrics}\label{sec:eval_metrics}
\subsubsection{Precision, Recall, and F1 Score:}
We use the reference closures computed in~\secref{sec:reference_closures} to evaluate our approach and the baseline methods quantitatively. Specifically, we compute \pr{} curves by varying the threshold~$\gamma$ on the number of inlier feature descriptor matches from our pipeline's RANSAC-based geometric validation stage. For the baseline methods, we generate their respective \pr{} curves by varying the key thresholding parameter described in their publications. In addition to the \pr{} curves, we report the average precision~(AP)~(area under the \pr{} curve), the maximum recall at 100\% precision~(R@1), and the maximum F1 score~(F1$_m$). We specifically choose to report R@1 as the maximum recall at 100\% precision to emphasize the need to avoid false loop closures in a SLAM pipeline that could lead to catastrophic failures~\citep{lowry2016tro}.

\subsubsection{Absolute Pose Error:}
We evaluate the effectiveness of our approach in correcting drift within a \slam{} pipeline through an offline \pg{} optimization using the g2o optimizer~\citep{kummerle2011icra}. We directly incorporate the detected loop closures between \maps{} as constraints in the \pg{}, using the initial transformation estimates from our pipeline.

We assess performance by computing the root mean square~(RMS) absolute pose error~(APE) in translation, with respect to ground-truth poses, both before and after the \pg{} optimization. To directly evaluate the accuracy of the detected loop closures and their alignments, we do not apply any robust kernel for outlier rejection during optimization.

We further evaluate the accuracy of the initial alignment estimate~($\T{m}{r}$) between loop-closed \maps{} using the RMS APE in translation and rotation.

\section{Experimental Evaluation}\label{sec:exp}

The primary focus of this work is an accurate and effective loop closure detection pipeline that works with various \lidar{} sensors, invariant to their scanning pattern, \fov{}, and resolution. We present our experiments to show the capabilities of our method and support our key claims. Our approach
(1) detects loop closures between \maps{} generated from various \lidar{} sensors with different scanning patterns, \fov{}s, and resolutions;
(2) performs multi-session loop closure detection and alignment with long-term revisits;
(3) works with handheld platforms having non-planar motion in the \lidar{} sensor frame;
(4) is robust against perceptual aliasing in environments with repetitive structures;
(5) provides a complete 3D rigid-body transform to align the detected loop closures;
(6) detects loop closures between sequences having minimal overlap, recorded with different \lidar{} sensor platforms, enabling cross-platform multi-map alignment.

\begin{figure*}[t]
	\begin{center}
		\def\svgwidth{0.915\linewidth}
		\captionsetup{type=figure}
		\input{figure_10.pdf_tex}
		\captionof{figure}{The \pr{} curves of \sota{} baselines and our approach for intra-session loop closure detection.}\label{fig:10}
	\end{center}
	\vspace{-15pt}
\end{figure*}

\begin{table*}[t!]
	\caption{Average precision~(AP), maximum recall at 100\% precision~(R@1), and maximum F1 scores~(F1$_m$) of \sota{} baselines and our approach for intra-session loop closure detection. Larger values indicate better performance. The best values are in bold, and the second-best values are underlined. The R@1 field is marked with ``-'' when the baseline cannot achieve a 100\% precision upon varying its corresponding thresholding parameter.}\label{tab:1}
	\vspace{5pt}
	\centering
	\begin{subtable}[t]{0.99\linewidth}
		\caption{\centering \tbf{\helipr{} dataset $\rightarrow$ \ouster{}}}\label{tab:1a}
		\small\sf\centering
		\setlength{\tabcolsep}{3.8pt}
		\begin{tabular}{lccc|ccc|ccc|ccc|ccc}
			\toprule
			\multirow{2}{*}[4pt]{Datasets} 	& \multicolumn{3}{c|}{Bridge01} 				& \multicolumn{3}{c|}{Bridge02} 				& \multicolumn{3}{c|}{Roundabout01} 			& \multicolumn{3}{c|}{Town02} 					& \multicolumn{3}{c}{Town03}                                                                                                                                                                           \\
			\midrule
			Metrics							& AP 			& R@1 			& F1$_m$ 		& AP 			& R@1 			& F1$_m$ 		& AP 			& R@1 			& F1$_m$ 		& AP 			& R@1 			& F1$_m$ 		& AP 			& R@1 			& F1$_m$ 		\\
			\midrule
			STD             				& 0.344 		& 0.049 		& 0.460 		& 0.300 		& - 			& 0.469 		& 0.316 		& 0.200 		& 0.420 		& 0.293 		& 0.059 		& 0.421 		& 0.373 		& 0.037 		& 0.538			\\
			STD-100		 					& 0.006 		& - 			& 0.048 		& 0.024 		& - 			& 0.121 		& 0.008 		& 0.005 		& 0.078 		& 0.005 		& 0.008 		& 0.076 		& 0.017 		& - 			& 0.133			\\
			BTC             				& \tbf{0.733} 	& 0.005 		& \tul{0.730} 	& 0.644 		& 0.009 		& 0.719 		& 0.627 		& 0.220 		& 0.689 		& \tbf{0.651} 	& 0.034 		& \tul{0.703} 	& 0.584 		& 0.022 		& 0.678			\\
			BTC-100		 					& 0.029 		& - 			& 0.171 		& 0.247 		& - 			& 0.393 		& 0.446 		& 0.031 		& 0.566 		& 0.098 		& - 			& 0.286 		& 0.013 		& - 			& 0.135			\\
			\solid{}        				& 0.136 		& 0.079 		& 0.211 		& 0.181 		& 0.009 		& 0.226 		& 0.072 		& 0.015 		& 0.121 		& 0.098 		& 0.008 		& 0.167 		& 0.106 		& - 			& 0.167			\\
			\logg3d{}						& 0.125 		& 0.007 		& 0.199 		& 0.127 		& 0.014 		& 0.188 		& 0.077 		& - 			& 0.162 		& 0.065 		& - 			& 0.128 		& 0.059 		& - 			& 0.112			\\
			\bevplace{}		 				& 0.347 		& 0.025 		& 0.445 		& 0.273 		& 0.009 		& 0.423 		& 0.167 		& 0.020 		& 0.227 		& 0.211 		& 0.068 		& 0.263 		& 0.168 		& - 			& 0.262			\\
			SC								& 0.186 		& - 			& 0.489 		& 0.166 		& - 			& 0.558 		& 0.217 		& 0.144 		& 0.412 		& 0.151 		& - 			& 0.382 		& 0.104 		& - 			& 0.351			\\
			MapClosure	    				& 0.662 		& \tul{0.219} 	& 0.720 		& \tbf{0.752} 	& \tul{0.470} 	& \tul{0.781} 	& \tul{0.731} 	& \tul{0.472} 	& \tul{0.788} 	& 0.568 		& \tul{0.119} 	& 0.690 		& \tul{0.694}	& \tul{0.343} 	& \tul{0.743}	\\
			\midrule
			\tbf{Ours}						& \tul{0.688} 	& \tbf{0.332} 	& \tbf{0.794} 	& \tul{0.685} 	& \tbf{0.614} 	& \tbf{0.797}	& \tbf{0.892} 	& \tbf{0.508} 	& \tbf{0.890} 	& \tul{0.607} 	& \tbf{0.424} 	& \tbf{0.750} 	& \tbf{0.711} 	& \tbf{0.388} 	& \tbf{0.749}	\\
			\bottomrule
		\end{tabular}
		\vspace{10pt}
	\end{subtable}
	\begin{subtable}[t]{0.99\linewidth}
		\caption{\centering \tbf{\helipr{} dataset $\rightarrow$ \aeva{}}}\label{tab:1b}
		\small\sf\centering
		\setlength{\tabcolsep}{3.8pt}
		\begin{tabular}{lccc|ccc|ccc|ccc|ccc}
			\toprule
			\multirow{2}{*}[4pt]{Datasets} 	& \multicolumn{3}{c|}{Bridge01} 							& \multicolumn{3}{c|}{Bridge02} 							& \multicolumn{3}{c|}{Roundabout01} 						& \multicolumn{3}{c|}{Town02} 								& \multicolumn{3}{c}{Town03}                                                                                                                                                                           \\
			\midrule
			Metrics							& AP 				& R@1 				& F1$_m$ 			& AP 				& R@1 				& F1$_m$ 			& AP 				& R@1 				& F1$_m$ 			& AP 				& R@1 				& F1$_m$ 			& AP 				& R@1 				& F1$_m$ 			\\
			\midrule
			STD								& 0.191 			& - 				& 0.376 			& 0.170 			& 0.005 			& 0.340 			& 0.087 			& - 				& 0.252 			& 0.082 			& 0.055 			& 0.177 			& 0.242 			& 0.010 			& 0.369 			\\
			STD-100		 					& 0.012 		& - 			& 0.080 		& 0.056 		& 0.005 		& 0.168 		& 0.021 		& 0.009 		& 0.082 		& 0.000 		& - 			& 0.058 		& 0.032 		& - 			& 0.113 		\\
			BTC								& 0.609 			& 0.076 			& 0.722 			& 0.511 			& 0.010 			& 0.679 			& 0.377 			& \tul{0.176} 		& 0.483 			& \tbf{0.259} 		& - 				& \tbf{0.429} 		& 0.371 			& 0.020 			& 0.494 			\\
			BTC-100		 					& 0.033 		& - 			& 0.192 		& 0.201 		& - 			& 0.392 		& 0.157 		& - 			& 0.367 		& 0.029 		& - 			& 0.150 		& 0.033 		& - 			& 0.180 		\\
			\solid{}						& 0.378 			& 0.166 			& 0.451 			& 0.156 			& - 				& 0.404 			& 0.211 			& 0.148 			& 0.277 			& 0.068 			& 0.027 			& 0.171 			& 0.235 			& - 				& 0.338 			\\
			\logg3d{}						& 0.041 			& 0.003 			& 0.103 			& 0.063 			& 0.010 			& 0.162 			& 0.038 			& - 				& 0.083 			& 0.030 			& - 				& 0.071 			& 0.058 			& 0.020 			& 0.106 			\\
			\bevplace{}		 				& 0.265 		& - 			& 0.414 		& 0.240 		& 0.005 		& 0.417 		& 0.082 		& 0.018 		& 0.139 		& 0.035 		& - 			& 0.098 		& 0.121 		& - 			& 0.168 		\\
			MapClosure						& \tbf{0.814} 		& \tbf{0.488} 		& \tbf{0.818} 		& \tbf{0.834} 		& \tbf{0.493} 		& \tbf{0.785} 		& \tul{0.478} 		& 0.139 			& \tul{0.607} 		& 0.196 			& \tbf{0.192} 		& 0.356 			& \tul{0.465} 		& \tbf{0.202} 		& \tul{0.584} 		\\
			\midrule
			\tbf{Ours}						& \tul{0.711} & \tul{0.424} & \tul{0.796} & \tul{0.683} & \tul{0.488} & \tul{0.765} &\tbf{0.584} 	& \tbf{0.259} & \tbf{0.654} & \tul{0.206} & \tul{0.110} & \tul{0.358} & \tbf{0.543} & \tul{0.172} & \tbf{0.642} \\
			\bottomrule
		\end{tabular}
		\vspace{10pt}
	\end{subtable}
	\begin{subtable}[t]{0.99\linewidth}
		\caption{\centering \tbf{\helipr{} dataset $\rightarrow$ \livox{}}}\label{tab:1c}
		\small\sf\centering
		\setlength{\tabcolsep}{3.8pt}
		\begin{tabular}{lccc|ccc|ccc|ccc|ccc}
			\toprule
			\multirow{2}{*}[4pt]{Datasets} 	& \multicolumn{3}{c|}{Bridge01}					& \multicolumn{3}{c|}{Bridge02} 				& \multicolumn{3}{c|}{Roundabout01} 			& \multicolumn{3}{c|}{Town02} 					& \multicolumn{3}{c}{Town03}                                                                                                                                                                           \\
			\midrule
			Metrics							& AP 			& R@1 			& F1$_m$		& AP 			& R@1 			& F1$_m$ 		& AP 			& R@1 			& F1$_m$ 		& AP 			& R@1 			& F1$_m$ 		& AP 			& R@1 			& F1$_m$ 		\\
			\midrule
			STD								& 0.056 		& - 			& 0.167 		& 0.073 		& 0.005 		& 0.178 		& 0.104 		& 0.024 		& 0.208 		& 0.062 		& 0.035 		& 0.171 		& 0.290 		& 0.111 		& 0.357 		\\
			STD-100		 					& 0.002 		& - 			& 0.025 		& 0.013 		& 0.010 		& 0.065 		& 0.015 		& 0.012 		& 0.056 		& 0.000 		& - 			& 0.028 		& 0.032 		& - 			& 0.107 		\\
			BTC								& 0.111 		& 0.040 		& 0.293 		& 0.211 		& - 			& 0.476 		& \tul{0.241} 	& \tul{0.146} 	& \tul{0.394} 	& \tul{0.133} 	& 0.035 		& \tbf{0.330} 	& 0.346 		& 0.022 		& 0.500 		\\
			BTC-100		 					& 0.011 		& - 			& 0.098 		& 0.105 		& - 			& 0.288 		& 0.050 		& 0.037 		& 0.175 		& 0.000 		& 0.018 		& 0.034 		& 0.010 		& 0.022 		& 0.079 		\\
			\solid{}						& 0.365 		& 0.052 		& 0.455 		& 0.400 		& 0.038 		& 0.462 		& 0.223 		& 0.024 		& 0.346 		& 0.064 		& - 			& 0.098 		& 0.212 		& - 			& 0.358 		\\
			\logg3d{}						& 0.013 		& - 			& 0.045 		& 0.020 		& - 			& 0.057 		& 0.026 		& 0.012 		& 0.059 		& 0.023 		& 0.018 		& 0.053 		& 0.036 		& - 			& 0.084 		\\
			\bevplace{}		 				& 0.205 		& - 			& 0.302 		& 0.200 		& 0.014 		& 0.321 		& 0.052 		& - 			& 0.112 		& 0.030 		& - 			& 0.092 		& 0.078 		& - 			& 0.144 		\\
			MapClosure						& \tul{0.512}	& \tbf{0.222} 	& \tul{0.609}	& \tbf{0.562} 	& \tul{0.229} 	& \tul{0.605} 	& 0.133 		& 0.073 		& 0.252 		& 0.102 		& \tbf{0.105} 	& 0.265 		& \tul{0.431} 	& \tbf{0.367} 	& \tul{0.576} 	\\
			\midrule
			\tbf{Ours}						& \tbf{0.549} 	& \tul{0.150} 	& \tbf{0.698} 	& \tul{0.543} 	& \tbf{0.300} 	& \tbf{0.688} 	& \tbf{0.278} 	& \tbf{0.158} 	& \tbf{0.442} 	& \tbf{0.141} 	& \tul{0.088} 	& \tul{0.269} 	& \tbf{0.464} & \tul{0.267} 	& \tbf{0.587} 	\\
			\bottomrule
		\end{tabular}
	\end{subtable}
\end{table*}

\begin{table*}[t!]
	\caption{Average precision~(AP), maximum recall at 100\% precision~(R@1), and maximum F1 scores~(F1$_m$) of \sota{} baselines and our approach for intra-session loop closure detection. Larger values indicate better performance. The best values are in bold, and the second-best values are underlined. The R@1 field is marked with ``-'' when the baseline cannot achieve a 100\% precision upon varying its corresponding thresholding parameter.}\label{tab:2}
	\vspace{5pt}
	\centering
	\begin{subtable}[t]{0.99\linewidth}
		\small\sf\centering
		\setlength{\tabcolsep}{3.8pt}
		\caption{\centering \tbf{\mulran{} dataset $\rightarrow$ Ouster OS1-64}}\label{tab:2a}
		\begin{tabular}{lccc|ccc|ccc|ccc|ccc}
			\toprule
			\multirow{2}{*}[4pt]{Datasets} 	& \multicolumn{3}{c|}{KAIST01} 					& \multicolumn{3}{c|}{KAIST02} 					& \multicolumn{3}{c|}{Riverside01} 				& \multicolumn{3}{c|}{Riverside02} 				& \multicolumn{3}{c}{Sejong01} 					\\
			\midrule
			Metrics							& AP 			& R@1 			& F1$_m$ 		& AP 			& R@1 			& F1$_m$ 		& AP 			& R@1 			& F1$_m$ 		& AP 			& R@1 			& F1$_m$ 		& AP 			& R@1 			& F1$_m$ 		\\
			\midrule
			STD             				& 0.545 		& 0.030 		& 0.655 		& 0.462 		& 0.258 		& 0.640 		& 0.329 		& - 			& 0.488			& 0.443 		& 0.067 		& 0.619 		& 0.016 		& - 			& 0.143 		\\
			STD-100		 					& 0.097 		& 0.007 		& 0.262 		& 0.080 		& - 			& 0.232 		& 0.203 		& 0.104 		& 0.318			& 0.103 		& 0.045 		& 0.202 		& 0.000 		& 0.000 		& 0.000 		\\
			BTC             				& \tul{0.680} 	& 0.007 		& 0.756 		& 0.605 		& 0.103 		& 0.736 		& \tbf{0.916} 	& 0.865 		& 0.927			& 0.645 		& - 			& 0.854 		& \tul{0.289} 	& \tul{0.364} 	& \tbf{0.909} 	\\
			BTC-100							& 0.293 		& 0.007 		& 0.502 		& 0.224 		& 0.010 		& 0.447 		& 0.615 		& 0.292 		& 0.706			& 0.456 		& 0.214 		& 0.609 		& 0.000 		& 0.000 		& 0.000 		\\
			\solid{}        				& 0.356 		& 0.104 		& 0.462 		& 0.457 		& 0.433 		& 0.604 		& 0.505 		& 0.271 		& 0.582			& 0.471 		& 0.326 		& 0.558 		& 0.010 		& - 			& 0.133 		\\
			\logg3d{}						& 0.246 		& 0.022 		& 0.362 		& 0.290 		& - 			& 0.353 		& 0.337 		& 0.021 		& 0.412			& 0.225 		& 0.022 		& 0.319 		& 0.001 		& - 			& 0.018 		\\
			\bevplace{}						& 0.387 		& 0.111 		& 0.421 		& 0.418 		& 0.237 		& 0.464 		& 0.490 		& 0.062 		& 0.556			& 0.428 		& 0.168 		& 0.451 		& 0.195 		& 0.091 		& 0.444 		\\
			SC              				& 0.176 		& - 			& 0.550 		& 0.169 		& 0.402 		& 0.636 		& 0.240 		& 0.417 		& 0.684			& 0.190 		& 0.404 		& 0.587 		& 0.162 		& 0.182 		& 0.424 		\\
			MapClosure						& 0.671 		& \tbf{0.437} 	& \tul{0.777} 	& \tul{0.790} 	& \tul{0.505} 	& \tul{0.828} 	& \tul{0.902} 	& \tul{0.885} 	& \tbf{0.946}	& \tul{0.831} 	& \tbf{0.708} 	& \tul{0.876} 	& 0.083 		& 0.182 		& 0.421 		\\
			\midrule
			\tbf{Ours}						& \tbf{0.733} 	& \tul{0.400} 	& \tbf{0.838} 	& \tbf{0.852} 	& \tbf{0.588} 	& \tbf{0.876} 	& 0.885 		& \tbf{0.896} 	& \tul{0.945}	& \tbf{0.843} 	& \tul{0.629} 	& \tbf{0.877} 	& \tbf{0.364} & \tbf{0.546} & \tul{0.706} 		\\
			\bottomrule\end{tabular}
		\vspace{10pt}
	\end{subtable}
	\begin{subtable}[t]{0.99\linewidth}
		\caption{\centering \tbf{\nclt{} dataset $\rightarrow$ \velodyne{}}; \tbf{\car{} dataset $\rightarrow$ Ouster OS1-128}; \tbf{\backpack{} dataset $\rightarrow$ Hesai Pandar-128}}\label{tab:2b}
		\small\sf\centering
		\setlength{\tabcolsep}{3.8pt}
		\begin{tabular}{lccc|ccc|ccc|ccc}
			\toprule
			\multirow{2}{*}[4pt]{Datasets} 	& \multicolumn{3}{c|}{\nclt{}~2012-01-08} 		& \multicolumn{3}{c|}{\nclt{}~2013-04-05} 		& \multicolumn{3}{c|}{\car{}} 					& \multicolumn{3}{c}{\backpack{}} \\
			\midrule
			Metrics							& AP 			& R@1 			& F1$_m$ 		& AP 			& R@1 			& F1$_m$ 		& AP 			& R@1 			& F1$_m$ 		& AP 			& R@1 			& F1$_m$ 		\\
			\midrule
			STD          					& \tul{0.429} 	& \tul{0.286} 	& 0.621 		& 0.219 		& \tul{0.417} 	& \tul{0.588} 	& 0.219 		& 0.068 		& 0.344 		& 0.261 		& 0.139 		& 0.323 		\\
			STD-100		 					& 0.000 		& 0.018 		& 0.035 		& 0.000 		& 0.000 		& 0.000 		& 0.013 		& 0.016 		& 0.042 		& 0.000 		& - 			& 0.053 		\\
			BTC          					& 0.333 		& - 			& \tul{0.654} 	& 0.104 		& 0.083 		& 0.333			& 0.470 		& 0.036 		& 0.607 		& 0.302 		& \tul{0.236} 	& 0.383 		\\
			BTC-100		 					& 0.319 		& - 			& 0.544 		& 0.000 		& 0.083 		& 0.154 		& 0.331 		& 0.112 		& 0.448 		& 0.000 		& 0.083 		& 0.154 		\\
			\solid{}     					& 0.092 		& 0.018 		& 0.153 		& 0.092 		& 0.018 		& 0.153 		& 0.078 		& 0.012 		& 0.120 		& 0.216 		& 0.014 		& 0.365 		\\
			\logg3d{}						& 0.070 		& - 			& 0.161 		& 0.019 		& - 			& 0.060			& 0.014 		& - 			& 0.045 		& 0.215 		& - 			& 0.355 		\\
			\bevplace{}		 				& 0.154 		& - 			& 0.274 		& 0.101 		& 0.167 		& 0.286			& 0.196 		& 0.028 		& 0.297 		& \tbf{0.370} 	& 0.125 		& \tul{0.398} 	\\
			SC           					& 0.144 		& - 			& 0.495 		& 0.031 		& - 			& 0.444			& 0.168 		& - 			& 0.426 		& 0.253 		& 0.083 		& 0.358 		\\
			MapClosure						& 0.321 		& 0.089 		& 0.500 		& \tul{0.307} 	& - 			& 0.571			& \tul{0.899} 	& \tul{0.272} 	& \tul{0.906} 	& 0.169 		& 0.181 		& 0.306 		\\
			\midrule
			\tbf{Ours} 						& \tbf{0.703} 	& \tbf{0.321} 	& \tbf{0.796} 	& \tbf{0.506} 	& \tbf{0.500} 	& \tbf{0.727} 	& \tbf{0.929} 	& \tbf{0.404} 	& \tbf{0.909} 	& \tul{0.361} 	& \tbf{0.361} 	& \tbf{0.540} 	\\
			\bottomrule
		\end{tabular}
	\end{subtable}
	\vspace{-5pt}
\end{table*}

\subsection{Intra-Session Loop Closure Detection}
In this experiment, we evaluate the performance of our approach on intra-session loop closure detection. We compare against several \sota{} baselines by converting their detected closures to \map{}-level closures, as described in~\secref{sec:reference_closures_baselines}. In addition, we include a comparison with our previous conference publication, MapClosure~\citep{gupta2024icra}, which forms the foundation of this work. We also evaluate the \map{}-based baselines STD and BTC using the same 100\,m travel displacement-based \maps{} as in our approach, referred to as STD-100 and BTC-100. However, we do not evaluate the scan-based baselines using similar \maps{} since scan-based methods like \scancontext{} and \solid{} depend on a \emph{single scan center} as the reference frame, which local maps inherently lack. Handling large lateral shifts would require sampling many artificial scan centers, changing the methods’ intended use.

The quantitative results in~\tabref{tab:1} and~\tabref{tab:2} show the effectiveness of our approach across multiple public and in-house datasets. Our method achieves the highest average precision in fifteen of the twenty-four sequences and ranks second in eight of the remaining nine. In terms of recall, it achieves the best performance at 100\% precision in fifteen sequences and the second-best in nine others. This balance between precision and recall is also reflected in the F1 scores, where our method ranks first in eighteen sequences and second in six others.

Our method provides a robust solution for loop closure detection within a \slam{} pipeline, where false positives must be minimized~\citep{blanco2013robotica,bailey2006ram,lowry2016tro}. The \pr{} curves in~\figref{fig:10} show that our approach~(red) consistently outperforms baselines across datasets with varied sensor setups. It maintains high precision over a broad range of recall values, eliminating the need for careful tuning of the inlier threshold in RANSAC or additional outlier rejection schemes in the \pg{} optimization.

The evaluations on the \helipr{} dataset demonstrate the versatility of our method across different \lidar{} sensors in varied urban environments. Our method consistently ranks among the top two methods across all metrics. In several sequences, our current approach ranks second only to MapClosure. However, the \pr{} curves in~\figref{fig:10} reveal a key advantage: our method maintains a higher precision than MapClosure, ensuring that detected loop closures remain accurate.

The \helipr{} Bridge sequences, which exhibit strong perceptual aliasing, highlight the effectiveness of our pruning strategy. While overall metrics in~\tabref{tab:1} may appear comparable between our method and MapClosure, our \pr{} curves in the first row of~\figref{fig:10} terminate earlier at a higher precision value than MapClosure, which also holds in comparison to other baselines. This behavior reflects our deliberate choice to prioritize precision over recall in perceptually ambiguous scenes. By doing so, we improve robustness, reduce sensitivity to the inliers threshold~($\gamma$), and lower the risk of false loop closures. The Town sequences present additional challenges with narrow alleyways, wide boulevards, and numerous dynamic objects such as pedestrians and cars. Even under these conditions, our approach ranks among the top two methods, showing its resilience in complex urban environments.

Additionally, the results on \helipr{} Bridge sequences provide insight into how the baselines perform under different levels of dynamic activity in the scene. The Bridge01 sequence, recorded at night, contains relatively few dynamic objects, whereas Bridge02, recorded during the day, includes substantially more dynamic participants. Since both sequences capture the same driving environment, we can directly compare their results to assess how dynamic objects affect loop closure detection. Our method performs consistently in terms of average precision and F1 score, highlighting its robustness to such dynamic elements. As expected, scan-based methods also show little to no variation in the metrics across these sequences, since individual scans are not significantly affected by dynamic objects due to the short temporal window of a single \lidar{} scan.

The ground alignment strategy introduced in~\secref{sec:ground_alignment} further strengthens performance on datasets with non-planar \lidar{} motion, such as the \nclt{} dataset and our self-recorded \backpack{} sequence. As shown in~\tabref{tab:2}, our method outperforms \scancontext{}, \bevplace{}, and MapClosure, which assume planar motion for the \bev{} projection. We achieve the best values across all three metrics, except for average precision on the \backpack{} dataset, where the performance gap to \bevplace{} remains marginal. Consistent \bev{} projection onto the local \ground{} during revisits, even under varying pitch and roll, drives this performance. On the \car{} dataset, recorded in hilly terrain with vegetation and elevation changes, our method achieves almost 100\% average precision, with a near-perfect \pr{} curve, confirming its robustness to complex \ground{} variations.

The results also confirm the importance of \maps{} for extracting meaningful structural information for place recognition and loop closure detection. Among the baselines, only STD, BTC, and MapClosure perform competitively, and all rely on \maps{}. Nonetheless, our evaluation of STD and BTC with the same \maps{} as our pipeline, referred to as STD-100 and BTC-100, shows that our method's advantage does not stem solely from larger \map{} sizes, but from the full algorithmic design.

In contrast, scan-based global descriptor approaches such as \scancontext{} and \solid{} fail to achieve consistent performance. \scancontext{}, designed for rotating \lidar{} sensors, performs poorly in the \helipr{} \ouster{} and \mulran{} sequences due to its strong dependence on viewpoint. \solid{}, designed to handle \lidars{} with different \fov{}s, performs better on the \helipr{} \aeva{} and \livox{} sequences but still falls short of \map{}-based methods. These trends highlight the inherent difficulty in constructing global descriptors for single \lidar{} scans.

The weak performance of \logg3d{} further illustrates this limitation. Without spatio-temporal aggregation or dimensionality reduction, \logg3d{} remains highly sensitive to the point density within each scan. Trained exclusively on sequences from the \mulran{} dataset with Ouster OS1-64, it performs well only on that dataset and generalizes poorly elsewhere. It fails to reach 100\% precision even at low recall values in most sequences, underscoring the need to retrain such methods for each sensor setup.

\bevplace{}, also a learning-based method, performs more robustly than \logg3d{} does due to its density-preserving \bev{} projection. However, it struggles on \nclt{} and \backpack{} sequences because of its assumption of planar motion for \bev{} projection.

Overall, these experiments confirm our core claim: our method delivers state-of-the-art performance in intra-session loop closure detection across datasets and \lidar{} platforms. By leveraging \maps{} as the core representation, our pipeline balances precision and recall, remains robust under perceptual aliasing, adapts to non-planar motion, and generalizes across different sensor characteristics, all without requiring changes to the pipeline parameters.

\subsection{Inter-Session and Inter-\lidar{} Loop Closure Detection}
In this experiment, we evaluate the performance of our approach for inter-session and inter-\lidar{} loop closure detection and compare it against \sota{} baselines introduced in~\secref{sec:baselines}. We select sequences spanning revisit intervals from a few weeks to several years to test the robustness under diverse temporal conditions.

Our pipeline architecture remains fundamentally the same in this scenario, with only minor modifications. We store the ground-alignment transformations for each \map{} together with the HBST database from the reference sessions. For a given query session, we match the ORB features against the reference database without adding any features from the query session to the database. Then, using the ground-alignment transforms of the corresponding \map{} from the reference session and the query \map{}, we compute a complete $\SE{3}$ pose estimate to align the loop closure between the two sessions.

We first evaluate inter-session loop closures using the same \lidar{} sensor. The \pr{} curves in~\figref{fig:11} and the metrics reported in~\tabref{tab:3} reveal trends consistent with the intra-session results. Our method consistently achieves the highest recall at 100\% precision across different \lidars{} and revisit periods, confirming its reliability in challenging conditions. The \pr{} curves highlight this robustness, as they terminate earlier but maintain high precision. Our method also ranks among the top two in average precision and maximum F1 scores, with BTC being the only baseline matching our overall performance. Other baselines often fail to reach 100\% precision or achieve it only at very low recall, except in the \mulran{} KAIST01–KAIST03 sequences.

\begin{figure*}
	\begin{center}
		\def\svgwidth{0.99\linewidth}
		\captionsetup{type=figure}
		\input{figure_11.pdf_tex}
		\captionof{figure}{The \pr{} curves of \sota{} baselines and our approach for inter-session loop closure detection.}\label{fig:11}
	\end{center}
\end{figure*}

\begin{table*}[h!]
	\caption{Average precision~(AP), maximum recall at 100\% precision~(R@1), and maximum F1 scores~(F1$_m$) of \sota{} baselines and our approach for inter-session loop closure detection between sequences recorded with the same \lidar{} sensor. Larger values indicate better performance. Best values are in bold, and the second-best values are underlined. The R@1 field is marked with ``-'' when the baseline cannot achieve a 100\% precision upon varying its corresponding thresholding parameter.}
	\label{tab:3}
	\vspace{0.5cm}
	\begin{subtable}[]{0.99\linewidth}
		\small\sf\centering
		\setlength{\tabcolsep}{3.5pt}
		\caption{\centering Conventional spinning \lidar{} sensors}\label{tab:3a}
		\begin{tabular}{lccc|ccc|ccc|ccc|ccc}
			\toprule
			Datasets			& \multicolumn{6}{c|}{\mulran{}} 																						& \multicolumn{3}{c|}{\nclt{}} 								& \multicolumn{6}{c}{\helipr{}} 																						\\
			\midrule
			Revisit Period		& \multicolumn{6}{c|}{$\approx$ 2 months} 																				& \multicolumn{3}{c|}{$>$1 year} 							& \multicolumn{3}{c|}{2 weeks} 								& \multicolumn{3}{c}{4 weeks} 								\\
			\midrule
			Reference Seq.		& \multicolumn{3}{c|}{KAIST01} 								& \multicolumn{3}{c|}{Sejong02} 							& \multicolumn{3}{c|}{2012-01-08} 							& \multicolumn{3}{c|}{Bridge02} 							& \multicolumn{3}{c}{Town01}								\\
			Query Seq.     		& \multicolumn{3}{c|}{KAIST03} 								& \multicolumn{3}{c|}{Sejong03} 							& \multicolumn{3}{c|}{2013-04-05} 							& \multicolumn{3}{c|}{Bridge03} 							& \multicolumn{3}{c}{Town03}								\\
			\midrule
			\lidar{}			& \multicolumn{3}{c|}{Ouster OS1-64} 						& \multicolumn{3}{c|}{Ouster OS1-64}						& \multicolumn{3}{c|}{\velodyne{}} 							& \multicolumn{3}{c|}{\ouster{}}							& \multicolumn{3}{c}{\ouster{}}								\\
			\midrule
			Metrics				& AP 				& R@1 			& F1$_m$ 				& AP 				& R@1 				& F1$_m$ 			& AP 				& R@1 				& F1$_m$ 			& AP 				& R@1 				& F1$_m$ 			& AP 				& R@1 				& F1$_m$ 			\\
			\midrule
			STD					& 0.451 			& 0.223 		& 0.529 				& 0.095 			& - 				& 0.224 			& 0.297 			& 0.241 			& 0.408 			& 0.220 			& - 				& 0.396 			& 0.266 			& 0.084 			& 0.360 			\\
			BTC					& \tul{0.635} 		& 0.309 		& \tul{0.740} 			& \tbf{0.378} 		& \tul{0.005} 		& \tbf{0.526} 		& \tbf{0.465} 		& \tul{0.311} 		& \tul{0.605} 		& \tbf{0.490} 		& 0.001 			& \tbf{0.647} 		& \tul{0.461} 		& \tul{0.259} 		& \tul{0.636} 		\\
			\solid{}			& 0.472 			& 0.085 		& 0.435 				& 0.022 			& - 				& 0.055				& 0.153 			& - 				& 0.307 			& 0.157 			& 0.004 			& 0.223 			& 0.239 			& 0.024 			& 0.276 			\\
			\logg3d{}			& 0.342 			& 0.027 		& 0.424 				& 0.017 			& - 				& 0.069 			& 0.179 			& - 				& 0.312 			& 0.152 			& \tul{0.005} 		& 0.205 			& 0.149 			& 0.002 			& 0.222 			\\
			\bevplace{}			& 0.547 			& 0.276 		& 0.575					& 0.053 			& -					& 0.149 			& 0.238 			& 0.017 			& 0.308				& 0.335 			& 0.002 			& 0.468 			& 0.316				& 0.025 			& 0.350				\\
			SC					& 0.160 			& \tul{0.445}	& 0.643 				& 0.039 			& - 				& 0.140 			& 0.252 			& - 				& 0.426 			& 0.205 			& - 				& 0.500 			& 0.247 			& - 				& 0.479 			\\
			\midrule
			\tbf{Ours}			& \tbf{0.725} 		& \tbf{0.633} 	& \tbf{0.835}			& \tul{0.203} 		& \tbf{0.037} 		& \tul{0.346}		& \tul{0.464} 		& \tbf{0.411} 		& \tbf{0.637}		& \tul{0.472} 		& \tbf{0.300} 		& \tul{0.638} 		& \tbf{0.497} 		& \tbf{0.437} 		& \tbf{0.664}		\\
			\bottomrule
		\end{tabular}
		\vspace{0.5cm}
	\end{subtable}
	\begin{subtable}[]{0.99\linewidth}
		\small\sf\centering
		\setlength{\tabcolsep}{3.5pt}
		\caption{\centering Unconventional hybrid solid-state and solid-state, non-spinning \lidar{} sensors}\label{tab:3b}
		\begin{tabular}{lccc|ccc|ccc|ccc}
			\toprule
			Datasets			& \multicolumn{12}{c}{\helipr{}} 																																																				\\
			\midrule
			Revisit Period  	& \multicolumn{6}{c|}{2 weeks} 																							& \multicolumn{6}{c}{4 weeks} 																							\\
			\midrule
			Reference Seq.		& \multicolumn{6}{c|}{Bridge02} 																						& \multicolumn{6}{c}{Town01} 																							\\
			Query Seq.			& \multicolumn{6}{c|}{Bridge03} 																						& \multicolumn{6}{c}{Town03} 																							\\
			\midrule
			\lidar{}			& \multicolumn{3}{c|}{\livox{}} 							& \multicolumn{3}{c|}{\aeva{}} 								& \multicolumn{3}{c|}{\livox{}} 							& \multicolumn{3}{c}{\aeva{}} 								\\
			\midrule
			Metrics				& AP 				& R@1 				& F1$_m$ 			& AP 				& R@1 				& F1$_m$ 			& AP 				& R@1 				& F1$_m$ 			& AP 				& R@1 				& F1$_m$ 			\\
			\midrule
			STD         		& 0.087 			& - 				& 0.168 			& 0.171 			& 0.001 			& 0.319 			& 0.238		 		& 0.066 			& 0.283 			& 0.226 			& \tul{0.088} 		& 0.270 			\\
			BTC            		& 0.161 			& 0.001 			& \tul{0.368} 		& \tul{0.391} 		& 0.008 			& \tbf{0.593} 		& \tul{0.295} 		& \tul{0.198} 		& \tul{0.442} 		& \tbf{0.415} 		& 0.068 			& \tbf{0.549} 		\\
			\solid{} 			& 0.151 			& 0.002 			& 0.209 			& 0.163 			& \tul{0.013} 		& 0.212				& 0.218 			& 0.058 			& 0.248 			& 0.224 			& 0.066				& 0.254 			\\
			\logg3d{}			& 0.039 			& - 				& 0.125 			& 0.065 			& - 				& 0.158 			& 0.114 			& 0.002				& 0.194 			& 0.152 			& 0.002 			& 0.249 			\\
			\bevplace{}			& \tul{0.207} 		& \tul{0.009} 		& 0.302				& 0.279 			& 0.002 			& 0.408				& 0.224 			& 0.078 			& 0.268				& 0.252 			& 0.054 			& 0.298				\\
			\midrule
			\tbf{Ours}			& \tbf{0.328} 		& \tbf{0.223} 		& \tbf{0.494} 		& \tbf{0.404} 		& \tbf{0.060} 	& \tul{0.580}			& \tbf{0.319} 		& \tbf{0.218} 		& \tbf{0.485}		& \tul{0.349} 	& \tbf{0.215} 			& \tul{0.519}		\\
			\bottomrule
		\end{tabular}
	\end{subtable}
\end{table*}

\begin{figure*}
	\begin{center}
		\def\svgwidth{0.99\linewidth}
		\captionsetup{type=figure}
		\input{figure_12.pdf_tex}
		\captionof{figure}{The \pr{} curves of \sota{} baselines and our approach for inter-\lidar{} loop closure detection.}\label{fig:12}
	\end{center}
	\vspace{-10pt}
\end{figure*}

\begin{table*}[t]
	\caption{Average precision~(AP), maximum recall at 100\% precision~(R@1), and maximum F1 scores~(F1$_m$) of \sota{} baselines and our approach for inter-session loop closure detection between sequences recorded with different \lidar{} sensors. Larger values indicate better performance. The best values are in bold, and the second-best values are underlined. The R@1 field is marked with ``-'' when the baseline cannot achieve a 100\% precision upon varying its corresponding thresholding parameter.}
	\label{tab:4}
	\vspace{0.5cm}
	\begin{subtable}[t]{0.99\linewidth}
		\small\sf\centering
		\setlength{\tabcolsep}{3.3pt}
		\caption{\centering}\label{tab:4a}
		\begin{tabular}{lccc|ccc|ccc|ccc|ccc}
			\toprule
			Datasets			& \multicolumn{15}{c}{\helipr{}} 																																																																	\\
			\midrule
			Revisit Interval	& \multicolumn{3}{c|}{2 weeks} 					& \multicolumn{3}{c|}{4 weeks} 					& \multicolumn{6}{c|}{$\approx$ 2 weeks} 														& \multicolumn{3}{c}{2 weeks} 					\\
			\midrule
			Reference Seq.		& \multicolumn{3}{c|}{Bridge02} 				& \multicolumn{3}{c|}{Bridge01} 				& \multicolumn{6}{c|}{Roundabout01} 															& \multicolumn{3}{c}{Town02} 					\\
			Query Seq.			& \multicolumn{3}{c|}{Bridge03} 				& \multicolumn{3}{c|}{Bridge03} 				& \multicolumn{6}{c|}{Roundabout02} 															& \multicolumn{3}{c}{Town03} 					\\
			\midrule
			Reference \lidar{}	& \multicolumn{3}{c|}{\ouster{}} 				& \multicolumn{3}{c|}{\ouster{}}				& \multicolumn{3}{c|}{\ouster{}} 				& \multicolumn{3}{c|}{\aeva{}} 					& \multicolumn{3}{c}{\ouster{}} 				\\
			Query \lidar{}		& \multicolumn{3}{c|}{\aeva{}} 					& \multicolumn{3}{c|}{\aeva{}}					& \multicolumn{3}{c|}{\aeva{}} 					& \multicolumn{3}{c|}{\livox{}} 				& \multicolumn{3}{c}{\aeva{}} 					\\
			\midrule
			Metrics				& AP 			& R@1 			& F1$_m$ 		& AP 			& R@1 			& F1$_m$ 		& AP 			& R@1 			& F1$_m$ 		& AP 			& R@1 			& F1$_m$ 		& AP 			& R@1 			& F1$_m$ 		\\
			\midrule
			STD          		& 0.086 		& 0.003 		& 0.208 		& 0.061 		& - 			& \tul{0.188} 	& 0.109 		& 0.003 		& 0.229 		& 0.019 		& - 			& 0.134 		& \tul{0.165} 	& 0.017 		& 0.250 		\\
			BTC             	& \tbf{0.398} 	& \tbf{0.033} 	& \tbf{0.592} 	& \tbf{0.352} 	& \tbf{0.004} 	& \tbf{0.520} 	& \tbf{0.499} 	& \tul{0.152} 	& \tbf{0.620} 	& \tbf{0.361} 	& 0.002 		& \tbf{0.504} 	& \tbf{0.478} 	& \tbf{0.243} 	& \tbf{0.612} 	\\
			\solid{} 			& 0.048 		& - 			& 0.097 		& 0.028 		& - 			& 0.072 		& 0.102 		& - 			& 0.228 		& \tul{0.099} 	& \tul{0.004} 	& \tul{0.198} 	& 0.105 		& - 			& \tul{0.251} 	\\
			\logg3d{}          	& 0.010 		& - 			& 0.084 		& 0.006 		& - 			& 0.061 		& 0.086 		& - 			& 0.232 		& 0.043 		& - 			& \tul{0.169} 	& 0.063 		& - 			& 0.204 		\\
			\bevplace{}			& 0.079 		& 0.003 		& 0.215			& 0.016 		& \tul{0.002} 	& 0.108			& 0.010 		& 0.001 		& 0.065			& 0.058 		& - 			& 0.188			& 0.055 		& - 			& 0.221			\\
			\midrule
			\tbf{Ours}			& \tul{0.156} 	& \tul{0.032} 	& \tul{0.280} 	& \tul{0.082} 	& - 			& 0.176 		& \tul{0.332} 	& \tbf{0.176} 	& \tul{0.502} 	& 0.067 		& \tbf{0.039} 	& 0.130			& 0.141 		& \tul{0.093} 	& \tul{0.251}	\\
			\bottomrule
		\end{tabular}
		\vspace{0.5cm}
	\end{subtable}
	\begin{subtable}[t]{0.99\linewidth}
		\small\sf\centering
		\setlength{\tabcolsep}{3.3pt}
		\caption{\centering}\label{tab:4_b}
		\begin{tabular}{lccc|ccc|ccc|ccc|ccc}
			\toprule
			Datasets			& \multicolumn{6}{c|}{\helipr{}} 																& \multicolumn{6}{c|}{\mulran{} $\times$ \helipr{}} 											& \multicolumn{3}{c}{Self-recorded} 			\\
			\midrule
			Revisit Interval	& \multicolumn{3}{c|}{2 weeks} 					& \multicolumn{3}{c|}{4 weeks} 					& \multicolumn{6}{c|}{$>$ 4 years} 																& \multicolumn{3}{c}{2 weeks} 					\\
			\midrule
			Reference Seq.		& \multicolumn{3}{c|}{Town02} 					& \multicolumn{3}{c|}{Town01} 					& \multicolumn{6}{c|}{\mulran{} KAIST01} 														& \multicolumn{3}{c}{\car{}}					\\
			Query Seq.      	& \multicolumn{3}{c|}{Town03} 					& \multicolumn{3}{c|}{Town03} 					& \multicolumn{6}{c|}{\helipr{} KAIST05} 														& \multicolumn{3}{c}{\backpack{}} 				\\
			\midrule
			Reference \lidar{}	& \multicolumn{3}{c|}{\livox{}} 				& \multicolumn{3}{c|}{\livox{}}					& \multicolumn{3}{c|}{Ouster OS1-64} 			& \multicolumn{3}{c|}{Ouster OS1-64} 			& \multicolumn{3}{c}{Ouster OS1-128} 			\\
			Query \lidar{}		& \multicolumn{3}{c|}{\aeva{}}					& \multicolumn{3}{c|}{\aeva{}}					& \multicolumn{3}{c|}{\ouster{}} 				& \multicolumn{3}{c|}{\aeva{}} 					& \multicolumn{3}{c}{Hesai Pandar-128} 			\\
			\midrule
			Metrics				& AP 			& R@1 			& F1$_m$ 		& AP 			& R@1 			& F1$_m$ 		& AP 			& R@1 			& F1$_m$ 		& AP 			& R@1 			& F1$_m$ 		& AP 			& R@1 			& F1$_m$ 		\\
			\midrule
			STD          		& 0.159 		& \tul{0.045} 	& 0.229 		& 0.159 		& \tul{0.034} 	& 0.226 		& 0.124 		& 0.014 		& 0.278 		& 0.108 		& - 			& \tul{0.271} 	& 0.012 		& \tul{0.012} 	& 0.052 		\\
			BTC             	& \tbf{0.325} 	& 0.002 		& \tbf{0.468} 	& \tbf{0.297} 	& 0.004 		& \tbf{0.437} 	& \tbf{0.585} 	& 0.100 		& \tbf{0.713} 	& \tbf{0.410} 	& \tbf{0.087} 	& \tbf{0.579} 	& \tbf{0.156} 	& \tbf{0.167} 	& \tul{0.286} 	\\
			\solid{} 			& \tul{0.163}	& - 			& 0.242 		& \tul{0.171} 	& 0.002 		& 0.250 		& 0.171 		& 0.002 		& 0.266 		& \tul{0.156} 	& - 			& 0.252 		& 0.005 		& - 			& 0.025 		\\
			\logg3d{}       	& 0.073 		& - 			& 0.187 		& 0.113 		& 0.002			& 0.186 		& 0.027 		& - 			& 0.162 		& 0.051 		& - 			& 0.194 		& 0.013 		& - 			& 0.050 		\\
			\bevplace{}			& 0.083 		& 0.003 		& 0.215			& 0.105 		& - 			& 0.236			& 0.151 		& 0.002 		& 0.359			& 0.140 		& - 			& \tul{0.303}	& 0.054 		& \tul{0.012} 	& 0.192			\\
			SC					& N/A 			& N/A 			& N/A 			& N/A 			& N/A 			& N/A 			& \tul{0.251} 	& \tbf{0.202} 	& \tul{0.432} 	& N/A 			& N/A 			& N/A 			& 0.031 		& - 			& 0.107 		\\
			\midrule
			\tbf{Ours}			& 0.145 		& \tbf{0.098} 	& \tul{0.259}	& 0.143 		& \tbf{0.044} 	& \tul{0.256}	& 0.229 		& \tul{0.106} 	& 0.377			& 0.091 		& \tul{0.054} 	& 0.200			& \tul{0.149} 	& \tbf{0.167} 	& \tbf{0.362} 	\\
			\bottomrule
		\end{tabular}
	\end{subtable}
\end{table*}

The Sejong02-Sejong03 sequences from the \mulran{} dataset were recorded as opposite-direction traversals on a long highway, which produces sparse structural geometry throughout the sequences. The restricted \fov{} of the dataset's \lidar{} further increases the difficulty of inter-session loop closure detection. As shown in~\tabref{tab:3a}, all baselines perform substantially worse on this pair of sequences, especially when we compare their results to those on the KAIST01-KAIST03 sequences from the same dataset, recorded with the same setup but in a dense urban environment.

The Bridge sequences from the \helipr{} dataset underscore the benefit of our self-similarity pruning strategy. Strong perceptual aliasing causes most baselines to either miss 100\% precision entirely or reach it only at negligible recall. Even BTC, which performs comparably to our method in most other scenarios, fails to maintain its high precision on the Bridge sequences because it lacks an explicit mechanism to handle perceptual aliasing. In contrast, our approach detects accurate loop closures at comparatively higher recall, as reflected in the clear separation of our \pr{} curves from the baselines.

Additionally, we achieve better performance across different \lidar{} sensors on the same sequences without changing any parameters in our pipeline. As shown in~\tabref{tab:3b}, the \aeva{} \lidar{} experiences a sharp drop in maximum recall at 100\% precision on the Bridge02-Bridge03 sequences. This drop arises from the overlap threshold used to define ground-truth loop closures: a few valid closures fall below the threshold and are therefore marked as false positives in ground-truth \maps{}, which lowers the maximum recall at 100\% precision. Nevertheless, the average precision and the corresponding \pr{} curve in~\figref{fig:11} demonstrate that our method still performs strongly, maintaining near-100\% precision across a substantially larger recall range.

On the \nclt{} dataset, which includes a revisit after about one year with a low-resolution \velodyne{} \lidar{}, our approach again achieves the best recall at 100\% precision and maximum F1 score. Only STD and BTC are the baselines that perform comparably in this scenario. This emphasizes the importance of having explicit knowledge about the \ground{} under non-planar \lidar{} motion.

These results also highlight the challenges that scan-based global descriptor methods such as \scancontext{}, \solid{}, and \logg3d{} face in long-term revisit scenarios. In such scenarios, significant lateral shifts, viewpoint changes during revisits, and scene changes can all alter the global descriptor substantially. In contrast, \map{}-based local descriptor methods like ours, STD, and BTC avoid these issues by aggregating multiple scans to reduce viewpoint dependence and by focusing on local regions of the scene that remain similar across revisits, rather than compressing the local context into a single global descriptor.

The results on the \helipr{} sequences further highlight the impact of the \lidar{} type on long-term place recognition. Even among the stronger baselines, we observe a noticeable drop in performance when switching from a~$360\degrees$-horizontal-\fov{} \ouster{} \lidar{} to the limited-horizontal-\fov{} \lidar{} sensors such as \aeva{} and \livox{}, even when evaluating the same pair of sequences.

We next evaluate inter-\lidar{} loop closure detection across multiple sessions. As shown in~\tabref{tab:4}, our approach ranks second overall to BTC but achieves the best recall at 100\% precision on five of ten sequences. The \pr{} curves in~\figref{fig:12} highlight the same trend: our method often detects fewer closures but preserves high precision, ensuring reliable results without extensive parameter tuning.

The performance of scan-based methods like \scancontext{}, \solid{}, \logg3d{}, and \bevplace{} is adversely affected by the difference in the scanning patterns, \fov{}s, and resolutions across \lidar{} sensors. The \map{}-based methods achieve better results in such cross-\lidar{} scenarios.

On the Roundabout01–Roundabout02, Town01–Town03, and Town02–Town03 sequences, our pipeline achieves the best recall at 100\% precision among all baselines, demonstrating robustness to opposite traversals and constrained sensor views. However, on Bridge01-Bridge03 and Bridge02-Bridge03 sequences, the feature-pruning strategy limits recall to very low values, leaving room for improvement.

Cross-dataset evaluations between \mulran{} and \helipr{} push this further, with revisit intervals of over 4 years and different \lidar{} sensors. These scenarios introduce structural changes in addition to sensor differences, making them highly challenging. Even so, our method achieves the second-best recall at 100\% precision, while all baselines, including ours, achieve lower recall overall.

Finally, in self-recorded datasets collected two weeks apart with platforms having different motion profiles and \lidars{}, our method performs comparably to BTC across all metrics, confirming its ability to generalize across diverse sensor setups.

Overall, these evaluations demonstrate that our approach effectively detects loop closures across multiple sessions for different \lidar{} sensors and revisit intervals, thereby supporting our second key claim. At the same time, they highlight opportunities for further research on the highly challenging problem of inter-\lidar{} loop closure detection.

\subsection{Analysis and Evaluation of the Ground Alignment Module}\label{sec:alignment_study}
In this section, we present both qualitative and quantitative evaluations of the ground alignment stage, which enhances the detection and alignment of loop closures. As we describe in~\secref{sec:ground_alignment}, we need this stage in scenarios where the \lidar{} experiences non-planar motion, allowing us to maintain a consistent reference plane for the \bev{} projection across revisits.

When we turn off the ground alignment module, our approach projects \maps{} onto the \xy{}, assuming planar motion. Consequently, using the RANSAC-based validation, it can align the \maps{} only along the \xy{}. We illustrate such a 2D alignment in the top-left image of~\figref{fig:13}, where two \maps{} appear well-aligned from a top-down view. However, this alignment can be misleading. In the top-right image of~\figref{fig:13}, a different perspective reveals an apparent misalignment in height and tilt between the two \maps{}, caused by non-planar \lidar{} motion.

By applying explicit ground alignment, we ensure that the \maps{} involved in a loop closure align correctly in 3D. This improvement enables our method to perform complete 3D global alignment rather than relying solely on a 2D assumption. The second row of~\figref{fig:13} shows a more accurate initial alignment between the \map{} pairs, even before any point cloud registration takes place.

We further validate this improvement through quantitative analysis. For each detected loop closure, we compute the overlap between the corresponding \maps{} after applying our initial transformation estimate~$\T{m}{r}$. We evaluate this on two sequences with non-planar \lidar{} motion: the 2013-04-05 sequence from the \nclt{} dataset~\citep{carlevaris-bianco2016ijrr} and the \backpack{} sequence recorded using our in-house setup. In~\tabref{tab:5}, we report the number of true positive loop closures with at least five inliers and the root mean square error of the overlap between aligned \maps{}, compared to the overlap between their reference counterparts generated with ground-truth poses.

As seen in~\tabref{tab:5}, the ground alignment stage significantly reduces the overlap error between \maps{}. This improvement occurs without applying any local point cloud registration. This advantage becomes evident in datasets with strong non-planar motion, such as the \backpack{} setup, where roll and pitch angles vary approximately between~-30$\degrees$ to~30$\degrees$. In such cases, the proposed ground alignment module enables our system to detect substantially more loop closures. This is because the \bev{} density images remain consistent across revisits when we project them onto the actual \ground{} in the environment rather than onto the \xy{} of the \map{} frame.

\begin{figure*}[ht]
	\begin{center}
		\def\svgwidth{0.99\linewidth}
		\captionsetup{type=figure}
		\input{figure_13.pdf_tex}
		\captionof{figure}{Visual comparison of loop-closed \map{} alignments with and without ground alignment. The top-down view~(xy) shows both \maps{} to be aligned in either case, as RANSAC operates in 2D. However, the side view~(xz) reveals significant misalignment without ground alignment due to uncorrected pitch and roll differences, as highlighted by the black ellipse. Applying ground alignment provides consistent 3D alignment by projecting the \maps{} onto a common physical \ground{}.}\label{fig:13}
	\end{center}
	\vspace{-5pt}
\end{figure*}

\begin{table}[th]
    \caption{Evaluation of the ground alignment strategy for loop closure detection and \map{} alignment. We report the number of correctly identified loop closures and the root mean square~(RMS) error in the overlap between aligned \maps{} using our pose estimate~$\T{m}{r}$ and the overlap between the reference \maps{}. We report results with and without the proposed ground alignment module.}\label{tab:5}
        \small\sf\centering
        \setlength{\tabcolsep}{4pt} 
        \begin{tabular}{lc|c|c|c}
            \toprule
            Dataset				                & \multicolumn{2}{c|}{\nclt{} 2013-04-05}   & \multicolumn{2}{c}{\backpack{}}   \\
                                                  \cmidrule(lr){2-3}                          \cmidrule(lr){4-5}
            Ground Alignment	                & OFF               & ON                    & OFF           & On                \\
            \midrule
            Correct Loop Closures~$\uparrow$	& \tbf{6}           & \tbf{6}               & 10            & \tbf{26}          \\
            RMS Error in Overlap~$\downarrow$	& 0.280	            & \tbf{0.110}           & 0.468         & \tbf{0.104}       \\
            Recall~$\uparrow$                   & \tbf{0.500}       & \tbf{0.500}           & 0.139         & \tbf{0.361}       \\
            \bottomrule
        \end{tabular}
        \vspace{-10pt}
    \end{table}

\begin{figure}[t]
	\begin{center}
		\def\svgwidth{0.95\linewidth}
		\captionsetup{type=figure}
		\centering
		\input{figure_14.pdf_tex}
		\captionof{figure}{Quantitative evaluation of the ground alignment accuracy. The azimuthal axis indicates the absolute initial misalignment magnitude between the \ground{} and \xy{} of the \maps{}. The radial axis (log scale) shows the absolute error of the predicted ground alignment magnitude, reported as mean values (dots) and standard deviation (bars), across different sequences and \lidars{} from the \mulran{} and \helipr{} datasets.}\label{fig:14}
	\end{center}
	\vspace{-5pt}
\end{figure}

We next evaluate the ability of our ground alignment module to correct misalignments between the \ground{} and the map's local \xy{}. To test this, we manually apply rotations in the range~(10$\degrees$\,-\,80$\degrees$) about ten random axes lying in the \xy{} of each reference \map{}. Our method then estimates the transform~$\T{g}{m}$ that realigns the \ground{} with the map's \xy{}. We perform this evaluation on the \mulran{} KAIST01 and Sejong01 sequences, as well as the \helipr{} Bridge01 sequence, using the \aeva{} and \livox{} \lidars{}. These sequences feature near-planar \lidar{} motion, so their reference \maps{} already have the \ground{} aligned with the \xy{}.

In~\figref{fig:14}, we report the mean and standard deviation of the absolute error in the predicted rotation magnitude, averaged across all ten axes for each \map{}, as a function of the applied misalignment. The ground alignment module reliably corrects misalignments up to~60$\degrees$, with a mean error below~1$\degrees$ and a standard deviation below~8$\degrees$. Even at initial misalignments of around~80$\degrees$, the mean absolute error typically remains below~10$\degrees$, except in the \livox{} sequence, where observed failures stem from the \livox{} sensor’s unusual reflective artifacts. These artifacts produce misleading, consistently oriented below-ground points that disrupt low-lying points sampling and PCA filtering under artificially large rotations. We would like to highlight that this is a sensor-specific quirk and not a flaw in the alignment method. However, since real-world misalignments rarely approach such extreme values, this level of performance is sufficient for practical use.

These results support our third claim that the proposed ground alignment module enables robust handling of handheld or non-planar \lidar{} motion and also provides a strong initial pose estimate for the 3D alignment of loop-closed \maps{}.

\subsection{Analysis and Evaluation of the Self-Similarity Pruning Strategy}\label{sec:pruning_study}

In this section, we evaluate the effectiveness of the self-similarity pruning strategy in detecting loop closures within environments characterized by highly repetitive structures. Specifically, we focus on the Bridge sequences from the \helipr{} dataset, which feature long bridges composed of repeating mechanical elements. These conditions present significant challenges due to perceptual aliasing, which can lead to false loop closures.

\figref{fig:15} shows a zoomed-in \pr{} curve for different self-similarity pruning thresholds based on the Hamming distance between ORB descriptors. The mark X on the curve represents the performance at an inlier threshold of 5, the default setting in our pipeline. Lower threshold values prune only strictly similar features, leading to higher recall values but a drop in precision, similar to the performance without pruning. Conversely, higher thresholds prune more features, leading to a drop in recall but maintaining high precision. The threshold value of 35 provides a desirable trade-off, maintaining a recall of around 0.6 with minimal loss in precision.

Without the pruning strategy, our method detects false loop closures in these Bridge sequences. We show one example at the bottom-left of~\figref{fig:17}, where a false match occurs with a near-perfect alignment between \maps{}. This alignment is supported by a sufficient number of inlier correspondences after a 2D RANSAC alignment of their \bev{} features~(right-hand side of~\figref{fig:17}), primarily caused by repetitive structures on the bridge. We successfully eliminate these matches between similar structures by applying the feature pruning strategy described in~\secref{sec:features}, thereby preventing the false loop closure.

\begin{figure}[t]
	\begin{center}
		\def\svgwidth{0.99\linewidth}
		\captionsetup{type=figure}
		\input{figure_15.pdf_tex}
		\captionof{figure}{Precision-recall curves for various choices of pruning thresholds in terms of the Hamming distance between ORB descriptors. We visualize a zoomed-in \pr{} curve for better differentiation between different curves.}\label{fig:15}
	\end{center}
\end{figure}

\begin{figure}[t]
	\begin{center}
		\def\svgwidth{0.99\linewidth}
		\captionsetup{type=figure}
		\input{figure_16.pdf_tex}
		\captionof{figure}{Precision-recall curves evaluating the impact of the pruning strategy for loop closure detection on the Bridge sequences from the \helipr{} dataset. These sequences are recorded in an environment with strong perceptual aliasing.}\label{fig:16}
	\end{center}
\end{figure}

\begin{table}[th!]
	\caption{The effect of feature pruning on loop closure accuracy, measured by RMS APE \wrt{} translation after \pg{} optimization. We report the results for three \lidar{} sensors across two sequences from the \helipr{} dataset.}\label{tab:6}
	\small\sf\centering
	\setlength{\tabcolsep}{4pt} 
	\begin{tabular}{lc|c|c|c|c|c}
		\toprule
		Metric 		& \multicolumn{6}{c}{RMS APE \wrt{} Translation~[m]~$\downarrow$} \\
					  \cmidrule(lr){2-7}
		Sequence 	& \multicolumn{3}{c|}{\helipr{} Bridge01} 		& \multicolumn{3}{c}{\helipr{} Bridge02} 		\\
					  \cmidrule(lr){2-4} 							  \cmidrule(lr){5-7}
		\lidar{} 	& Ouster 		& Aeva 			& Livox 		& Ouster 		& Aeva 			& Livox 		\\
		\midrule
		Odometry	& 178.8			& 124.4 		& 223.0 		& 52.4 			& 149.3 		& 114.9 		\\
		w/o Pruning & 331.1 		& 282.7 		& 695.9 		& 572.6 		& 314.1 		& 457.7 		\\
		w/ Pruning  & \textbf{23.9} & \textbf{28.7} & \textbf{51.0} & \textbf{16.6} & \textbf{23.5} & \textbf{56.4} \\
		\bottomrule
	\end{tabular}
\end{table}

We further compare the performance of our pipeline with and without feature pruning in~\figref{fig:16}. The \pr{} curves for both configurations, across the two Bridge sequences and three different \lidar{} sensors, show that our approach with pruning maintains high precision throughout rather than compromising on precision to achieve a better recall. In contrast, our approach without pruning achieves slightly higher recall but significantly drops precision.

\begin{figure*}[t]
	\begin{center}
		\def\svgwidth{0.99\linewidth}
		\captionsetup{type=figure}
		\input{figure_17.pdf_tex}
		\captionof{figure}{Visualization of perceptual aliasing in the \helipr{} Bridge sequence and the impact of feature pruning on loop closure detection. \textit{Left:} The repetitive semi-circular arches of the bridge~(top) lead to structurally similar \maps{}~(below) as highlighted by the yellow ellipses, causing a false loop closure. \textit{Right:} Feature correspondence plots before and after pruning. Without pruning, RANSAC mistakenly accepts many inlier correspondences~(green lines), resulting in a false positive. After pruning, only outlier matches~(red lines) remain, correctly preventing erroneous loop closure. Feature pruning effectively removes ambiguous, repetitive features that lead to perceptual aliasing.}\label{fig:17}
	\end{center}
\end{figure*}

\begin{table*}[h]
	\caption{Evaluation of the loop closure alignment accuracy. We report accuracy in terms of RMS APE in translation~(m) and rotation~($\degrees$) compared to the proxy ground-truth alignment obtained via Open3D point-to-point ICP. Best results are in \tbf{bold}.}\label{tab:7}
	{
	\begin{subtable}[t]{0.99\linewidth}
		\small\sf\centering
		\setlength{\tabcolsep}{3.3pt}
		\vspace{10pt}
		\caption{\centering We report the number of true loop closures and the accuracy of the initial pose estimates~($\T{m}{r}$) produced by our method. Results from KISS-Matcher~(KM) are also shown to contextualize the quality of our estimates; KISS-Matcher’s values correspond to only successful alignments.}\label{tab:7a}
		\begin{tabular}{l|c|c|c|c|c|c|c}
			\toprule
			\multirow{2}{*}[-0.3em]{Sequence} 	& \multirow{2}{*}[-0.3em]{\lidar{}} & \multicolumn{2}{c|}{\# Closures} & \multicolumn{2}{c|}{RMS APE~(translation)\,[m]} & \multicolumn{2}{c}{RMS APE~(rotation)\,[$\,\degrees$]} \\
			\cmidrule(lr){3-4}\cmidrule(lr){5-6}\cmidrule(lr){7-8}
			&     													& KM	& Ours	& KM	& Ours	& KM	& Ours	\\
			\midrule
			
			\multirow{3}{*}{\helipr{} Bridge01}		& \ouster{}		& 215 & \tbf{216} & \tbf{5.38 $\pm$ 5.34} & 5.57 $\pm$ 5.40 & \tbf{0.87 $\pm$ 0.83} & 1.80 $\pm$ 1.47 \\
													& \livox{}		& \tbf{122} & \tbf{122} & 15.68 $\pm$ 14.95 & \tbf{2.07 $\pm$ 1.24} & \tbf{2.09 $\pm$ 1.87} & 2.74 $\pm$ 1.91 \\
													& \aeva{}		& 213 & \tbf{214} & 1.65 $\pm$ 1.43 & \tbf{2.17 $\pm$ 1.50} & \tbf{1.06 $\pm$ 0.87} & 1.90 $\pm$ 1.44 \\
			\midrule
			\multirow{3}{*}{\helipr{} Town02}		& \ouster{}		& \tbf{42} & \tbf{42} & \tbf{0.30 $\pm$ 0.20} & 0.89 $\pm$ 0.44 & \tbf{0.31 $\pm$ 0.17} & 1.11 $\pm$ 0.84 \\
													& \livox{}		& \tbf{5} & \tbf{5} & \tbf{1.07 $\pm$ 0.62} & 1.46 $\pm$ 0.73 & \tbf{0.65 $\pm$ 0.26} & 0.82 $\pm$ 0.44 \\
													& \aeva{}		& \tbf{9} & \tbf{9} & \tbf{0.55 $\pm$ 0.23} & 1.31 $\pm$ 0.60 & \tbf{1.32 $\pm$ 0.96} & 1.62 $\pm$ 0.90 \\
			\midrule
			\multirow{3}{*}{\helipr{} Roundabout01}	& \ouster{}		& 125 & \tbf{139} & \tbf{0.33 $\pm$ 0.22} & 0.72 $\pm$ 0.33 & \tbf{0.33 $\pm$ 0.23} & 0.56 $\pm$ 0.33 \\
													& \livox{}		& 12 & \tbf{13} & \tbf{0.44 $\pm$ 0.27} & 2.52 $\pm$ 1.88 & \tbf{0.56 $\pm$ 0.40} & 2.04 $\pm$ 1.45 \\
													& \aeva{}		& 44 & \tbf{51} & \tbf{1.02 $\pm$ 0.74} & 1.63 $\pm$ 0.90 & \tbf{1.14 $\pm$ 0.83} & \tbf{1.14 $\pm$ 0.61} \\
			\midrule
			\car{}									& Ouster OS1-128 & 216 & \tbf{230} & \tbf{0.49 $\pm$ 0.40} & 0.84 $\pm$ 0.49 & \tbf{0.50 $\pm$ 0.36} & 1.15 $\pm$ 0.80 \\
			\midrule
			\nclt{} 2012-01-08						& \velodyne{}	& 14 & \tbf{26} & \tbf{1.06 $\pm$ 0.87} & 1.23 $\pm$ 0.85 & \tbf{1.06 $\pm$ 0.86} & 1.28 $\pm$ 0.97 \\
			\bottomrule
		\end{tabular}
		\vspace{0.75cm}
	\end{subtable}
	\begin{subtable}[t]{0.99\linewidth}
		\small\sf\centering
		\setlength{\tabcolsep}{3.3pt}
		\caption{\centering We refine the pose estimates from both methods using Open3D point-to-point ICP, and report the RMS APE in translation and rotation relative to the reference alignment. Both methods converge to similarly accurate solutions, demonstrating that our initial pose estimates are sufficiently precise to serve as reliable initialization for ICP-style refinement within a SLAM pipeline.}\label{tab:7b}
		\begin{tabular}{l|c|c|c|c|c}
			\toprule
			\multirow{2}{*}[-0.3em]{Sequence} 	& \multirow{2}{*}[-0.3em]{\lidar{}} & \multicolumn{2}{c|}{RMS APE~(translation)\,[m]} & \multicolumn{2}{c}{RMS APE~(rotation)\,[$\,\degrees$]} \\
			\cmidrule(lr){3-4}\cmidrule(lr){5-6}
			&     														& KM + ICP & Ours + ICP & KM + ICP & Ours + ICP \\
			\midrule
			
			\multirow{3}{*}{\helipr{} Bridge01}		& \ouster{}			& \tbf{5.05 $\pm$ 5.03} & \tbf{5.05 $\pm$ 5.05} & 1.12 $\pm$ 1.11 & \tbf{0.73 $\pm$ 0.73} \\
													& \livox{}			& 15.67 $\pm$ 14.98 & \tbf{0.47 $\pm$ 0.43} & 2.02 $\pm$ 1.90 & \tbf{0.70 $\pm$ 0.69} \\
													& \aeva{}			& 1.51 $\pm$ 1.41 & \tbf{0.60 $\pm$ 0.56} & 0.72 $\pm$ 0.70 & \tbf{0.40 $\pm$ 0.39} \\
			\midrule
			\multirow{3}{*}{\helipr{} Town02}		& \ouster{}			& \tbf{0.02 $\pm$ 0.01} & \tbf{0.02 $\pm$ 0.01} & \tbf{0.03 $\pm$ 0.01} & \tbf{0.03 $\pm$ 0.01} \\
													& \livox{}			& \tbf{0.04 $\pm$ 0.02} & \tbf{0.04 $\pm$ 0.02} & \tbf{0.02 $\pm$ 0.01} & \tbf{0.02 $\pm$ 0.01} \\
													& \aeva{}			& \tbf{0.02 $\pm$ 0.01} & \tbf{0.02 $\pm$ 0.01} & \tbf{0.03 $\pm$ 0.01} & \tbf{0.03 $\pm$ 0.01} \\
			\midrule
			\multirow{3}{*}{\helipr{} Roundabout01}	& \ouster{}			& \tbf{0.02 $\pm$ 0.01} & \tbf{0.02 $\pm$ 0.01} & \tbf{0.02 $\pm$ 0.01} & \tbf{0.02 $\pm$ 0.01} \\
													& \livox{}			& 0.23 $\pm$ 0.22 & \tbf{0.17 $\pm$ 0.16} & 0.45 $\pm$ 0.43 & \tbf{0.23 $\pm$ 0.22} \\
													& \aeva{}			& 0.48 $\pm$ 0.47 & \tbf{0.26 $\pm$ 0.25} & 0.49 $\pm$ 0.48 & \tbf{0.25 $\pm$ 0.25} \\
			\midrule
			\car{}									& Ouster OS1-128 	& 0.30 $\pm$ 0.29 & \tbf{0.28 $\pm$ 0.28} & \tbf{0.17 $\pm$ 0.17} & \tbf{0.17 $\pm$ 0.16} \\
			\midrule
			\nclt{} 2012-01-08						& \velodyne{}		& \tbf{0.63 $\pm$ 0.61} & 0.94 $\pm$ 0.89 & 0.65 $\pm$ 0.62 & \tbf{0.61 $\pm$ 0.57} \\
			\bottomrule
		\end{tabular}
	\end{subtable}
	}
\end{table*}

To underscore the importance of precision, we conduct a \pg{} optimization experiment using the loop closures detected under both settings. As shown in~\tabref{tab:6}, loop closures obtained with pruning lead to markedly better absolute trajectory estimates despite detecting fewer closures. In contrast, disabling pruning results in more false positives, which degrade the final trajectory estimate, even surpassing the error of odometry-only results.

Overall, this study supports our fourth claim that the self-similarity feature pruning strategy introduced in~\secref{sec:features} is essential for maintaining robustness in loop closure detection under high structural repetition. By prioritizing precision, our method avoids catastrophic failures in pose estimation, ultimately leading to more reliable mapping and localization.

\subsection{Evaluation of the 3D Alignment Estimates}\label{sec:pose_accuracy}
In this section, we evaluate the accuracy of the initial alignment~($\T{m}{r}$) for the loop closures detected by our approach across the \helipr{}, \nclt{}, and \car{} datasets. To obtain a reliable reference for evaluation, we refine the \map{} alignments using Open3D's~\citep{zhou2018arxiv} point-to-point ICP registration, initialized with the dataset ground-truth pose. We use these refined results strictly as a \emph{proxy ground-truth alignment}, since directly comparing against the dataset poses would conflate alignment error with odometry drift accumulated within the \maps{} that we generate using \lidar{} odometry.

In~\tabref{tab:7a}, we report the number of correctly identified loop closures and the RMS APE in translation and rotation for the pose estimates~($\T{m}{r}$) produced by our method. Our loop closure identification approach consistently yields initial alignments within a couple of meters of translation and a few degrees of rotation relative to the reference alignment. This performance holds across all evaluated \lidars{}, despite their differing scanning patterns, resolutions, and \fov{}s. It can thus be used with ICP for fine alignment.

We also compare our method to KISS-Matcher by~\cite{lim2025icra}~(KM), a state-of-the-art global point cloud registration method based on 3D features. For KISS-Matcher, we report the number of loop closures it successfully aligns~(\ie{}, more than five inliers) and compute the corresponding RMS error metrics with respect to the reference alignment. We include this comparison not as a competing baseline, but to contextualize the quality of our \emph{initial alignment}, and demonstrate how close our pose estimates are to those produced by a dedicated global registration system.

As shown in~\tabref{tab:7}, KISS-Matcher, a technique designed to globally register point clouds, achieves better alignment accuracy across most sequences. This is expected as KISS-Matcher computes 3D features from the 3D point cloud and estimates the pose using a maximally consistent correspondence set. In contrast, our method combines 2D feature-based pose estimates with the approximate ground-alignment estimates to get a complete 3D pose. Despite this fundamental difference, our RMS APE values remain within a few decimeters in translation and a couple of degrees in rotation compared to those of KISS-Matcher. Notably, KISS-Matcher performs worse than our method on the \helipr{} Bridge01 sequence because it lacks an explicit mechanism to handle perceptual aliasing during alignment.

To further demonstrate that our pose estimates are well-suited as initialization for a fine ICP-style alignment stage within a SLAM system, we explicitly run a point-to-point ICP refinement using Open3D, initialized with the pose estimates from our method as well as KISS-Matcher. The results in~\tabref{tab:7b} show that both methods converge to solutions that are similarly accurate, and close to the reference alignment.

Our refined results also outperform KISS-Matcher on several sequences. The apparent performance inversion between KISS-Matcher and our pose estimation accuracy before and after ICP refinement may seem counterintuitive at first. A plausible explanation lies in the non-linear least-squares optimization underlying the ICP algorithm, which is sensitive to initialization and can converge to different local minima depending on the initial pose estimate.

Furthermore, two \maps{} generated from the same location using \lidar{} odometry will generally not align perfectly, as they accumulate different amounts of drift and may contain artifacts due to imperfect scan deskewing. This could also lead to a different convergence for different initializations. It is also important to note that the discrepancy in performance between the two methods in~\tabref{tab:7b} is, in most cases, on the order of only a few centimeters. The overall trend is that both methods converge to similarly accurate solutions after ICP refinement; therefore, we make no claims about the superiority of one method over the other in terms of final alignment accuracy.

Overall, these results support our fifth claim that the proposed approach yields accurate and complete 3D rigid-body transforms that align detected loop closures and are sufficiently precise to serve as initial guesses for fine-grained ICP-style registration within a SLAM back-end.

\subsection{Multi-Map Alignment}

This final experiment evaluates our pipeline's ability to detect loop closures between the Riverside03 sequence and three KAIST sequences from the \mulran{} dataset. Although these trajectories share only minimal spatial overlap, it is sufficient to align the two scenes. These sequences also vary in temporal separation, with revisits ranging from the same day to a few months apart.

We first compare our method against baseline approaches using the \pr{} curves shown in~\figref{fig:18}. Our approach consistently outperforms the baselines, achieving the highest recall while maintaining 100\% precision across all three cases. Most other methods, except \scancontext{}, struggle in this challenging setting, with precision often falling below 50\%.

Next, we compute the optimized trajectory for each sequence using in-session loop closure constraints detected by our method, followed by an additional \pg{} optimization step to align these optimized sequences across sessions using the inter-session closures identified by our pipeline.
\begin{figure*}[h]
	\begin{center}
		\def\svgwidth{0.99\linewidth}
		\captionsetup{type=figure}
		\input{figure_18.pdf_tex}
		\captionof{figure}{Precision-recall curves of our approach and \sota{} baselines for loop closure detection across sequences from the \mulran{} dataset with limited overlap.}\label{fig:18}
	\end{center}
\end{figure*}

\begin{figure*}[h]
	\begin{center}
		\def\svgwidth{0.99\linewidth}
		\captionsetup{type=figure}
\begingroup%
  \makeatletter%
  \providecommand\color[2][]{%
    \errmessage{(Inkscape) Color is used for the text in Inkscape, but the package 'color.sty' is not loaded}%
    \renewcommand\color[2][]{}%
  }%
  \providecommand\transparent[1]{%
    \errmessage{(Inkscape) Transparency is used (non-zero) for the text in Inkscape, but the package 'transparent.sty' is not loaded}%
    \renewcommand\transparent[1]{}%
  }%
  \providecommand\rotatebox[2]{#2}%
  \newcommand*\fsize{\dimexpr\f@size pt\relax}%
  \newcommand*\lineheight[1]{\fontsize{\fsize}{#1\fsize}\selectfont}%
  \ifx\svgwidth\undefined%
    \setlength{\unitlength}{443.32450807bp}%
    \ifx\svgscale\undefined%
      \relax%
    \else%
      \setlength{\unitlength}{\unitlength * \real{\svgscale}}%
    \fi%
  \else%
    \setlength{\unitlength}{\svgwidth}%
  \fi%
  \global\let\svgwidth\undefined%
  \global\let\svgscale\undefined%
  \makeatother%
  \begin{picture}(1,0.42191828)%
    \lineheight{1}%
    \setlength\tabcolsep{0pt}%
    \put(0,0){\includegraphics[width=\unitlength,page=1]{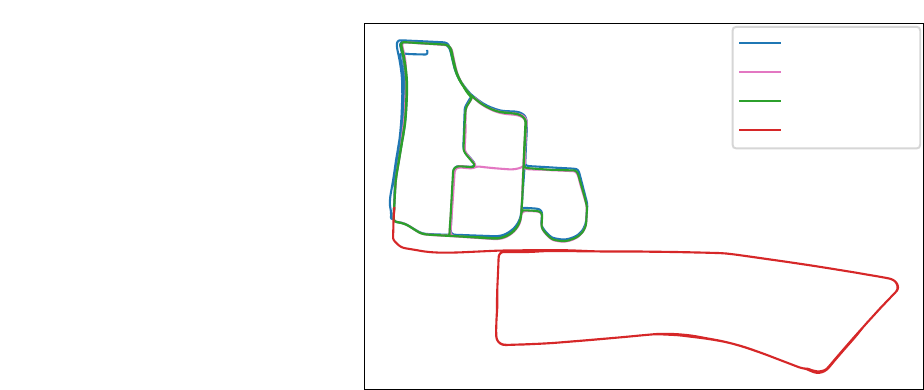}}%
    \put(0.85840944,0.36815109){\makebox(0,0)[lt]{\lineheight{1.25}\smash{\begin{tabular}[t]{l}KAIST01\end{tabular}}}}%
    \put(0.85840944,0.33681919){\makebox(0,0)[lt]{\lineheight{1.25}\smash{\begin{tabular}[t]{l}KAIST02\end{tabular}}}}%
    \put(0.85840944,0.30548724){\makebox(0,0)[lt]{\lineheight{1.25}\smash{\begin{tabular}[t]{l}KAIST03\end{tabular}}}}%
    \put(0.85839847,0.27394608){\makebox(0,0)[lt]{\lineheight{1.25}\smash{\begin{tabular}[t]{l}Riverside03\end{tabular}}}}%
    \put(0,0){\includegraphics[width=\unitlength,page=2]{figure_19.pdf}}%
    \put(0.16914289,0.40625081){\makebox(0,0)[t]{\lineheight{1.25}\smash{\begin{tabular}[t]{c}aligned \map{}s through loop closure\end{tabular}}}}%
    \put(0.69716883,0.40624531){\makebox(0,0)[t]{\lineheight{1.25}\smash{\begin{tabular}[t]{c}aligned trajectories\end{tabular}}}}%
    \put(0,0){\includegraphics[width=\unitlength,page=3]{figure_19.pdf}}%
  \end{picture}%
\endgroup%

		\captionof{figure}{We align three KAIST sequences to the Riverside03 sequence from the \mulran{} dataset using our loop closure pipeline. Despite the minimal spatial overlap, highlighted by the dashed rectangle, our approach accurately detects loop closures and aligns the corresponding \maps{}.}\label{fig:19}
	\end{center}
	\vspace{-15pt}
\end{figure*}

We show in~\figref{fig:19} the final aligned trajectories and corresponding \maps{}, illustrating the successful alignment in the overlapping region. We report the precision, recall, and F1 scores for the inter-session loop closures in~\tabref{tab:8}, using the default parameters detailed in~\secref{sec:baselines}. Despite the limited overlap, our method achieves perfect precision in all cases.

To assess the quality of the predicted alignment transformation~$\T{m}{r}$, we compute the RMS error of the overlap between aligned \maps{}, compared to the overlap between their reference counterparts generated with ground-truth poses. The last row of~\tabref{tab:8} demonstrates that our alignment closely matches ground-truth performance.

Additionally, we perform a similar cross-session multi-map alignment between the two self-recorded datasets, \car{} and \backpack{}, captured with different \lidar{} sensors mounted on separate mobile platforms. These sequences exhibit varying motion dynamics and sensor configurations but still contain overlapping regions. We show the resulting aligned trajectories and detected loop closures in~\figref{fig:1}.

\begin{table}[t]
	\caption{Precision, recall, and F1 score for multi-session loop closures between the Riverside03 and KAIST sequences from the \mulran{} dataset. We report the RMS error in the overlap between aligned \maps{} using our pose estimate~$\T{m}{r}$ and the overlap between the reference \maps{}.}\label{tab:8}
	\small\sf\centering
	\setlength{\tabcolsep}{5pt} 
	\begin{tabular}{lc|c|c}
		\toprule
		Reference Session                   & \multicolumn{3}{c}{Riverside03} 	\\
		\cmidrule(lr){2-4}
		Query Session                       & KAIST01	& KAIST02  	& KAIST03 	\\
		\midrule
		Revisit Interval                    & 2 Months  & Same Day 	& 1 Week  	\\
		\midrule
        Precision~$\uparrow$               	& 1.0       & 1.0      	& 1.0     	\\
		Recall~$\uparrow$                  	& 0.120     & 0.108    	& 0.150   	\\
		F1~$\uparrow$                      	& 0.214     & 0.195    	& 0.261   	\\
		\midrule
		RMS Error in Overlap~$\downarrow$	& 0.043		& 0.055   	& 0.085   	\\
		\bottomrule
	\end{tabular}
	\vspace{-10pt}
\end{table}

This experiment supports our final claim that our approach can successfully detect and align loop closures in challenging scenarios with little overlap. This capability is critical for tasks such as multi-robot map alignment, collaborative mapping, and long-term change detection.

\subsection{Runtime Evaluation of Our Approach}
In this additional experiment, we evaluate the runtime of our approach. We run all the experiments on an Intel i9-10980XE @ 3.00\,GHz CPU with 64\,GB RAM\@. We implement our approach in C++ with a single-thread design. We report the runtime of different components of our pipeline in~\tabref{tab:9}, including the mean and standard deviations for performing ground alignment for a \map{}, detecting loop closures using \bev{} density images, and validating detected loop closures through RANSAC\@. We also provide the average number of scans in each \map{}. The time required to generate each \map{} is not reported, as it directly corresponds to the \lidar{} odometry algorithm used,~\ie{},~\kiss{}.

Additionally, in~\tabref{tab:10}, we compare the average runtimes of all baselines in terms of frames processed per second~(FPS) across three sequences with dense, 360\,$\degrees$ \fov{} \lidars{}. \scancontext{} achieves the highest processing speed on all sequences, followed by \solid{} and our method. Notably, the FPS values for STD, BTC, and our method include the runtime of \kiss{} for scan registration, which generally makes them slower than scan-based methods such as \solid{} and \scancontext{}, which do not require registration. The learning-based methods, \logg3d{} and \bevplace{}, are the slowest, often running below the typical \lidar{} frame rate of 10\,FPS\@.

\begin{table*}[t]
	\caption{Runtime evaluation of our approach. We report the mean and standard deviation of the execution times for each key component of our pipeline.}\label{tab:9}
	\small\sf\centering
	\begin{tabular}{lc|c|c|c}
		\toprule
								& 							& \multicolumn{3}{c}{Execution Time (mean $\pm$ standard deviation) [ms]} 	\\
															  \cmidrule(lr){3-5}
 		Sequence				& Avg. Scans per Local Map 	& Ground Alignment 	& Loop Closure Detection	& Loop Closure Validation	\\
		\midrule
		\mulran{} Sejong01     	& 119	 					& 4.27 $\pm$ 1.59	& 17.74 $\pm$ 10.85			& 0.31 $\pm$ 0.11 			\\
		\helipr{} Bridge01     	& 96  	 					& 7.44 $\pm$ 4.02  	& 32.11 $\pm$ 21.21 		& 0.34 $\pm$ 0.20 			\\
		\nclt{}~2012-01-08     	& 664 	 					& 14.94 $\pm$ 7.21	& 63.79 $\pm$ 50.49 		& 0.38 $\pm$ 0.31 			\\
		\car{}                 	& 137 	 					& 10.38 $\pm$ 6.04  & 50.96 $\pm$ 28.40    		& 0.35 $\pm$ 0.33			\\
		\bottomrule
	\end{tabular}
\end{table*}

\begin{table*}[t]
	\caption{Average runtime evaluation of all the baselines. We report the average number of frames processed per second~(FPS) for each baseline on three sequences with dense, 360\,$\degrees$ \fov{} \lidars{}.}\label{tab:10}
	\small\sf\centering
	\begin{tabular}{lc|c|c}
		\toprule
		Sequence		& \mulran{} KAIST01	& \helipr{} Bridge01	& \nclt{} 2012-01-08	\\
						\cmidrule{2-4}
		\lidar{}		& Ouster OS1-64		& \ouster{}				& \velodyne{}			\\
		\midrule
		STD				& 24  				& 9 					& 13 					\\
		BTC				& 13  				& 8 					& 17 					\\
		\scancontext{}	& \tbf{144} 		& \tbf{101} 			& \tbf{111}				\\
		\solid{}		& \tul{35}			& \tul{24}				& \tul{40}				\\
		\logg3d{}		& 12  				& 7						& 15					\\
		\bevplace{}		& 7   				& 5						& 5						\\
		\midrule
		\tbf{Ours}		& 27  				& 15					& 29					\\
		\bottomrule
	\end{tabular}
	\vspace{-5pt}
\end{table*}

Overall, our approach achieves a favorable balance between speed and effectiveness, outperforming most conventional baselines while remaining significantly faster than other \map{}-based and learning-based methods, consistently maintaining processing speeds above the sensor frame rate.

In summary, the experiments presented above support all our claims and showcase the applicability of our approach across various scenarios. The first experiment shows the robustness of our approach to various scanning patterns, \fov{}s, and resolutions of the \lidar{} sensor. The second experiment illustrates the applicability of our approach to detect loop closures between multiple sessions with short-term and long-term revisits through the same environment. The two subsequent experimental evaluations support the newly proposed modules in our loop closure pipeline. We show that our approach can be used on \lidar{} platforms with non-planar motion and is also robust towards perceptual aliasing. The fifth experiment confirms that our pipeline produces accurate and complete 3D pose estimates for aligning loop closures. We further demonstrate that our approach can align multiple trajectories with minimal physical scene overlap. Finally, we show that our pipeline can efficiently operate above sensor frame rates, even on dense 3D \lidar{} sensors.
\section{Limitations}\label{sec:limitations}
Our approach demonstrates robustness across multiple datasets, but it has certain limitations. The pipeline relies on locally consistent \lidar{} odometry to generate \maps{}, which constitutes its most critical limitation. When the scan matching algorithm produces degenerate maps, loop closure performance also degrades. Incorporating more robust odometry that fuses inertial or visual data with \lidar{} scans can help alleviate this limitation.

The method also depends on consistent \ground{} detection for \bev{} projection, which may not hold in cluttered indoor, forested, or off-road scenarios. In such environments, approaches tailored to specific settings, such as ForestLPR~\citep{shen2025cvpr}, perform better. For aerial data, the density-preserving \bev{} projection can undersample vertical structures; a maximum-height \bev{} may provide a stronger alternative.

Although our method consistently performs better than existing baselines in intra-session and inter-session scenarios, inter-\lidar{} scenarios remain challenging. Opposite-direction revisits with low-\fov{} \lidar{} sensors, such as \livox{} and \aeva{}, are particularly difficult due to limited scene overlap, even after aggregation into \maps{}.

In summary, our method performs robustly across diverse urban datasets but is constrained by odometry quality, assumptions of a detectable \ground{}, and reduced recall in the inter-\lidar{} loop closure detection. Addressing these limitations will broaden its applicability to more diverse environments.

\section{Conclusion}\label{sec:conclusion}

In this paper, we present a novel and robust approach to detect loop closures for \lidar{}-based \slam{} and provide 3D alignments between the detected closures. Our method leverages a density-preserving bird's-eye-view projection of \maps{} generated from local odometry estimates. We use the local \ground{} as a common reference plane across revisits and align the \maps{} such that the \ground{} coincides with the \xy{} of the \map{}'s reference frame. This ensures a consistent \bev{} representation across diverse mobile platforms with varying \lidar{} motion profiles. We extract ORB feature descriptors on these \bev{} projections and apply a self-similarity pruning strategy to reduce incorrect closures in repetitive environments.

We implement and extensively evaluate our approach on several public and self-recorded datasets featuring different \lidar{} sensors mounted on various mobile platforms. We compare our method with \sota{} baselines and demonstrate its effectiveness in detecting loop closures in diverse urban settings across \lidar{} sensors with different scanning patterns and fields of view. Furthermore, we illustrate the capability of our approach to detect long-term loop closures across multiple sessions. Finally, our runtime evaluations indicate the real-time feasibility of our approach.

To facilitate further research on place recognition and loop closure detection in \lidar{}-\slam{}, we release our software as open-source.

\begin{dci}
  The author(s) declared no potential conflicts of interest with respect to the research, authorship, and/or publication of this article.
\end{dci}

\begin{funding}
This work has partially been funded 
  by the Deutsche Forschungsgemeinschaft (DFG, German Research Foundation) under Germany's Excellence Strategy, EXC-2070 -- 390732324 -- PhenoRob,
  by the Deutsche Forschungsgemeinschaft (DFG, German Research Foundation) under STA~1051/5-1 within the FOR 5351~(AID4Crops),
  by the European Union’s Horizon Europe research and innovation programme under grant agreement No~101070405~(DigiForest),
  and by the German Federal Ministry of Research, Technology and Space~(BMFTR) under the Robotics Institute Germany~(RIG).
\end{funding}

\bibliographystyle{SageH}
\bibliography{new}

\end{document}